\documentclass[10pt,twocolumn,letterpaper]{article}

\usepackage[pagenumbers]{cvpr} % To force page numbers, e.g. for an arXiv version

\usepackage{caption}
\usepackage{makecell}
\usepackage{tikz}
\usetikzlibrary{spy, calc}
\usepackage{soul} 
\usepackage{xcolor} 
\usepackage{algorithm}
\usepackage{algorithmic}
\usepackage{multirow}
\usepackage{graphicx}
\usepackage{float}
\usepackage[resetlabels,labeled]{multibib}
\usepackage[accsupp]{axessibility}  % Improves PDF readability for those with disabilities.

\definecolor{cvprblue}{rgb}{0.21,0.49,0.74}
\usepackage[pagebackref,breaklinks,colorlinks,allcolors=cvprblue]{hyperref}
\usepackage{cancel}
\def\paperID{15326} % *** Enter the Paper ID here
\def\confName{CVPR}
\def\confYear{2025}

\title{Identity-preserving Distillation Sampling by Fixed-Point Iterator}

\newcommand*\samethanks[1][\value{footnote}]{\footnotemark[#1]}
\author{
SeonHwa Kim$^1$ \quad Jiwon Kim$^1$ \quad Soobin Park$^2$ \quad Donghoon Ahn$^1$ \quad Jiwon Kang$^1$\\
\quad Seungryong Kim$^3$\thanks{Corresponding author.} \quad Kyong Hwan Jin$^1$\samethanks \quad Eunju Cha$^2$\samethanks\\
$^1$Korea University \quad\qquad $^2$Sookmyung Women's University \quad\qquad $^3$KAIST\\
{\tt\small \{sunkim0062, jwonkim, dhahn99, jiwon7258, kyong\_jin\}@korea.ac.kr,} \\
{\tt\small \{psb1219j, eunju.cha\}@sookmyung.ac.kr, seungryong.kim@kaist.ac.kr}
}

\begin{document}
\maketitle
\begin{abstract}
Score distillation sampling (SDS) demonstrates a powerful capability for text-conditioned 2D image and 3D object generation by distilling the knowledge from learned score functions. However, SDS often suffers from blurriness caused by noisy gradients. When SDS meets the image editing, such degradations can be reduced by adjusting bias shifts using reference pairs, but the de-biasing techniques are still corrupted by erroneous gradients. To this end, we introduce Identity-preserving Distillation Sampling (IDS), which compensates for the gradient leading to undesired changes in the results. Based on the analysis that these errors come from the text-conditioned scores, a new regularization technique, called fixed-point iterative regularization (FPR), is proposed to modify the score itself, driving the preservation of the identity even including poses and structures. Thanks to a self-correction by FPR, the proposed method provides clear and unambiguous representations corresponding to the given prompts in image-to-image editing and editable neural radiance field (NeRF). The structural consistency between the source and the edited data is obviously maintained compared to other state-of-the-art methods. Our code is \url{https://github.com/shhh0620/IDS}
\end{abstract}

\section{Introduction}
\label{sec:intro}

\begin{figure}[t]
    \centering
    % First subfigure
    \begin{subfigure}[t]{\columnwidth}
        \centering
        \includegraphics[width=1\textwidth]{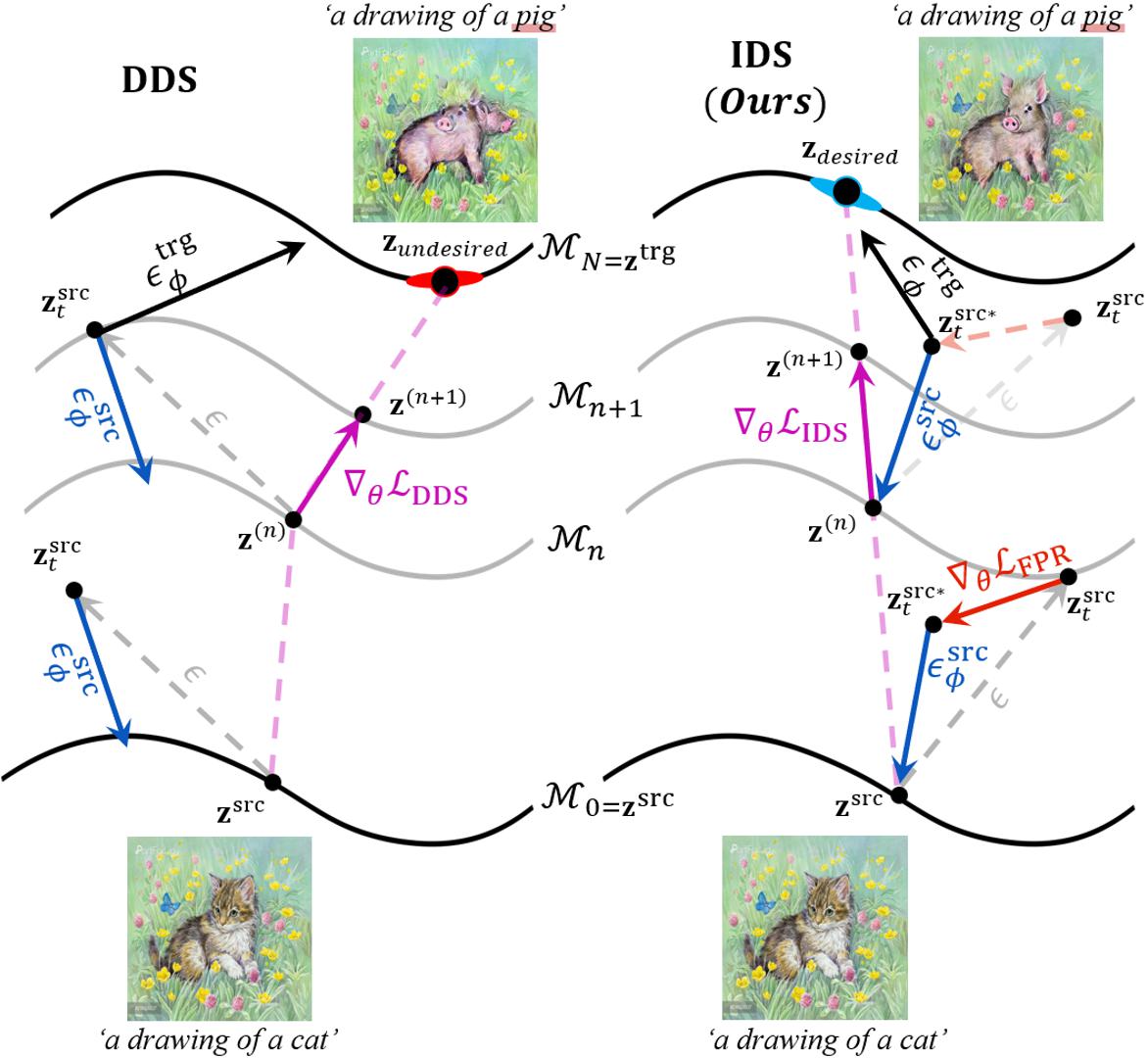} % Replace with your image file
        % \caption{Manifold structures of DDS \& \textit{Ours}}
        % \label{fig:subfig1}
    \end{subfigure}%
    % \hfill
    % Second subfigure
    % \begin{subfigure}[t]{\columnwidth}
    %     \centering
    %     \includegraphics[width=0.7\textwidth]{Fig./Algorithm/imgs/algorithm_ours_4.png} % Replace with your image file
    %     \caption{Ours}
    %     \label{fig:subfig2}
    % \end{subfigure}
    
    % \begin{subfigure}[t]{\columnwidth}
    %     \centering
    %     \includegraphics[width=.8\textwidth]{Fig./Algorithm/imgs/teasor_5.png} % Replace with your image file
    %     \caption{Flow of IDS}
    %     \label{fig:subfig3}
    % \end{subfigure}
    \caption{\textbf{Trace of guided updating} from source to target images using delta denoising score (DDS) and identity-preserving distillation sampling (IDS). DDS moves a gradient of score function toward $\mathcal M_\mathbf{z}$ manifold directed by stochastic direction $\epsilon$. In contrast, IDS moves a gradient with a corrected direction by a fixed-point regularization.}
    \vspace{-6pt}
    \label{fig:algorithm}
\end{figure}

\vspace{-6pt}
% diffusion general story
Diffusion models ~\cite{ho2020denoising,song2020score,dhariwal2021diffusion,ho2021classifier,rombach2022high} have shown powerful representations on text-to-image (T2I) generative tasks. With the advance of classifier guidance (CG) and classifier-free guidance (CFG) paradigms~\cite{dhariwal2021diffusion,ho2021classifier,hong2023improving,ahn2024self}, diffusion models improve the quality of generated samples \cite{ho2020denoising,song2020score}. Such high-quality image generators can be easily extended to image editing by simply modifying forward/reverse iterations~\cite{mengsdedit}, applying CFG with a target prompt~\cite{brooks2023instructpix2pix,hertzprompt} or interchanging attention layers \cite{tumanyan2023plug}. 

\begin{figure*}[!t]
        \centering
        \includegraphics[width=0.85\textwidth]{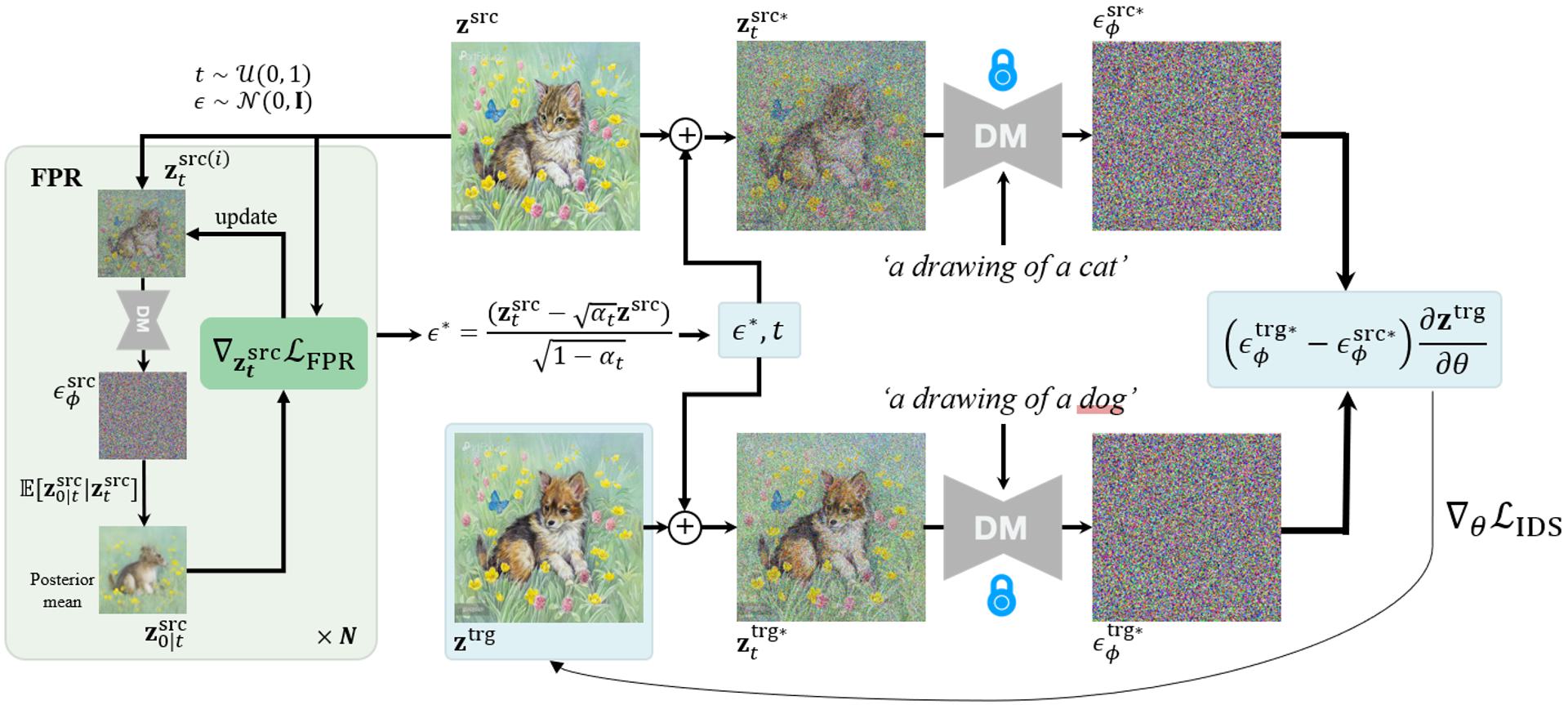} % Replace with your image file
    \caption{\textbf{Flowchart of IDS.} The backbone of our algorithm employs DDS \cite{hertz2023delta} framework to distill score function into a target image. Our fixed-point regularization (FPR) obtains a guided noise, $\epsilon^*$, from iterative updates using posterior mean computed by Tweedie’s formula. When distilling the score function to a target image, the guided noise is updated while maintaining the identity of the source.}
    \label{fig:teasor}
\end{figure*}

% editing by DM strong and weakness with SDS
Recently, Delta Denoising Score (DDS)  \cite{hertz2023delta} is proposed to edit a source image by distilling the rich generative prior of T2I diffusion models. It is based on the analysis of Score Distillation Sampling (SDS)~\cite{poole2022dreamfusion}, originally developed to optimize a parametric generator such as Neural Radiance Fields (NeRFs)~\cite{mildenhall2021nerf} by exploiting the learned score of the diffusion models. Even though SDS offers remarkable performance in synthesizing 3D scenes, noisy gradients from stochastic perturbations lead to significantly over-saturated results that faithfully follow the given text prompts. In the context of image editing, text prompts do not often include information about the identity of the source image, such as the background, the object's pose, or the structure of the content, which should be retained during updates. Thus, DDS is designed to resolve such blurriness by erasing gradients of non-text-aligned features from SDS gradients. There is no explicit procedure to preserve the source's identity in DDS updates because the fine gradient may provide the conserved identity. However, this cannot be guaranteed if many variations in the structure are possible, such as editing the image of a cat into a pig, as shown in Fig.~\ref{fig:algorithm}. To alleviate this problem, Contrastive Denoising Score (CDS)~\cite{nam2024contrastive} and Posterior Distillation Sampling (PDS)~\cite{koo2024posterior} are introduced to maximize the mutual information of the source image and edited image. Although such algorithms rely heavily on text prompts, the algorithms have yet to analyze the inherent error caused by text-conditioned scores.

%However, the inevitable loss of identity caused by the text-conditioned score is not addressed. This problem arises from the fundamental nature of a text prompt, which can describe an infinite number of images, including the source image itself. 

%With DDS, however, even more severe errors occur, as the identity of the original image, such as the structure of the content and the background, is not explicitly taken into account. The edited images therefore often have deformed shapes or colors. 

% 

%To alleviate this problem, Contrastive Denoising Score (CDS) \cite{nam2024contrastive} is suggested to incorporate the contrastive learning scheme \cite{chen2020simple, park2020contrastive} within DDS framework to maximize the mutual information of the source image and edited image at the latent feature level. \cite{koo2024posterior} introduced a variant of DDS loss to explicitly match the stochastic latents during optimization. 
%However, these algorithms still rely on noisy scores caused by overfitting to text prompts.
%However, the text-conditioned score corresponds to the gradient from the generated latent by a forward diffusion process from the source image to the specific image represented by the corresponding text prompt. 
%Since a single text prompt can describe infinite possible images, the score conditioned by the source text prompt ($y^{src}$) is not guaranteed to always lead to the source image. This implies that the identity of the original image is inevitably lost during editing.

To this end, we investigate the underlying meaning of text-conditioned score. The gradient maps the stochastic latent, generated by applying the forward diffusion process to the given image, to one of the possible images described by the prompt, including the original image.
%Even if the desired direction points towards the original image, the actual gradient often deviates from the desired direction.
Simply, the score obtained from the latent of the source image (`source latent') and the source prompt can be a gradient to another image represented by the identical text. %, which deviates from the desired direction to the source image. 
Based on this interpretation, the accumulation of misaligned directions causes the loss of the source's identity, leading to structural changes in the result with DDS, as shown in Fig.~\ref{fig:algorithm}.

To address this issue, we propose a novel score distillation sampling to effectively preserve the identity of the original image by self-correcting the misaligned gradients, called \textbf{I}dentity-preserving \textbf{D}istillation \textbf{S}ampling (\textbf{IDS}).
The key insight is that if the score is precisely adjusted to the source image, the conditional expectation of the source image given the source latent contains meaningful information that should be preserved during the editing.
This conditional expectation corresponds to the posterior mean computed by Tweedie’s formula using the learned score~\cite{efron2011tweedie, chungdiffusion}. The source latent is iteratively updated to make the posterior mean similar to the source image. This procedure, named a fixed-point iterative regularization (FPR), results in the aligned score with the source that provides reliable gradients for editing, as illustrated in Fig.~\ref{fig:algorithm}.
%In particular, a fixed-point iterative regularization (FPR) is introduced to provide reliable gradients to DDS \cite{hertz2023delta} inspired by a fixed-point iteration~\cite{parikh2014proximal}.
%by simply applying fixed-point iterator~\cite{parikh2014proximal} to the text-conditioned score obtained from a source latent ($\mathbf{z}_{t}^{\text{src}}$) and a source text prompt ($y^{src}$) with respect to a posterior mean image.
%The source latent ($\mathbf{z}_{t}^{\text{src}}$) is iteratively updated to extract the optimal injection noise ($\epsilon^*$), which is used to obtain the gradient for the DDS update.
Following IDS update is performed using guided noise extracted from the refined source latent, rather than random Gaussian noise. This further ensures the identity preservation.
%Our FPR improves DDS updates by refining gradient noises toward reliable pre-trained manifolds in iterative updates of a reference latent($\hat z_t$)
%to minimize the difference between the given image and the posterior mean, resulting in the optimal injection noise ($\epsilon^*$).
%is then extracted from the updates source latent, which produces the score adjusted to the source image.
%The DDS loss is then computed to produce the transformed image based on the correct score. The procedure for fixing the misalignment of the gradient to the source image is dubbed fixed-point iterative regularization (FPR).
%This self-correction approach is simple but effective in refining the gradient toward reliable pre-trained manifolds, while maintaining the identity. 
%Our FPR improves DDS updates by refining gradient noises toward reliable pre-trained manifolds in iterative updates of a reference latent($\hat z_t$).{\color{red} what is the specific defacts in DDS?}
% \add{A paragraph about the experimental results}
Our method demonstrated superior results compared with baselines in two tasks: editing images by prompts and editing NeRF.

In summary, our main contributions are as follows:
\begin{itemize}

    \item We obtain reliable gradients for the score distillation function by a fixed-point iterator with respect to posterior means. The iterator corrects the text-conditioned score, guiding SDS gradients toward reliable pre-trained manifolds.

    \item Our fixed-point regularization preserves the identities of sources such as structures and poses in edited targets for 2D and 3D editing. Such preservation is well demonstrated in qualitative and quantitative results. %both IoU and background PSNR between source and target.

    % \item Our method achieves convergent target images which earn the highest score by user study and GPT compared with baselines. %other score distillation methods for text-guided editing.

\end{itemize}

% \item We obtain reliable gradients for score distillation function by a fixed-point iterator with respect to posterior means. The iterator guides SDS gradient toward reliable pre-trained manifolds.

% \item Our fixed-point iterator is universally applicable to score distillation sampling approaches such as SDS \cite{poole2022dreamfusion}, DDS \cite{hertz2023delta} and CDS \cite{nam2024contrastive} as a self-correcting regularizer. 

\section{Related works}
\label{sec:rel_works}
%\subsection{Controls in Diffusion models}
% LDM 의 cross-attention : text prompt guidance를 제공
% 큰 틀에서, text2image, text-based or guidance-based editing 모두 diffusion model에 guidance를 주는것. 그것을 training-free상황에서 진행하는것이 SDEdit 이고, trainable 하게 image를 바꾸는것이 SDS계열임
%Guidance-based diffusion models provide an crucial direction for diffusion models to generate higher-quality images.
%CG\cite{dhariwal2021diffusion} and CFG\cite{ho2021classifier} are prominent approaches in this domain, balancing sample quality and diversity by incorporating guidance during both denoising and sampling. These approaches have inspired further developments to enhance diffusion models' quality and adaptability across diverse conditioning settings, broadly encompassing tasks like text-to-image synthesis and text- or guidance-based editing, all of which provide targeted guidance to diffusion models.
% CG CFG 는 sampling 과정에서의 guidance를 제공.
\subsection{Image Editing with Diffusion Models}
% sdeedit 은 strok등의 인풋을 적당한 노이즈로 corrupt 시켜 guidance로 함.
With the great success of image generation using diffusion models, the pre-trained diffusion models have been recently employed for image editing tasks, 
%Recently, various diffusion models have been employed for image editing tasks, 
demonstrating significant advancements in the quality and flexibility of generated edits \cite{mengsdedit, hertzprompt, tumanyan2023plug, brooks2023instructpix2pix, nam2024contrastive, koo2024posterior}. 
Stochastic Differential Editing (SDEdit) \cite{mengsdedit} is a pioneering work in which the source image was modified by adding noise and solving reverse stochastic differential equations.
Thanks to the text-conditional Latent Diffusion Model (LDM), a.k.a. Stable Diffusion \cite{rombach2022high}, text-driven editing approaches have been introduced. Specifically, the text embedding was injected through the cross-attention layer of the model for image editing and translation, while retaining the structure of the original image \cite{hertzprompt, tumanyan2023plug}. The editing was further controlled by rescaling the attention of the specific word \cite{hertzprompt} or by manipulating the self-attention features \cite{tumanyan2023plug}.
These approaches provide greater control by balancing fidelity between the edited prompt and the source image without the need for model training, fine-tuning, additional data, or optimization. However, the current DDIM-based inversion \cite{song2020denoising} can lead to unsatisfactory reconstructions for real images, and the cross-attention bottleneck limits its effectiveness for broader edits. Crafting suitable prompts also remains challenging for complex compositions. 

\subsection{Score distillation sampling}
Score Distillation Sampling (SDS)~\cite{poole2022dreamfusion} enables text-driven 3D synthesis by leveraging probability density distillation loss to distill knowledge from 2D diffusion models, allowing high-quality 3D scene generation based on textual prompts without 3D training data. However, SDS has limitations, often producing oversaturated and overly smooth 3D models, and lacking diversity across initializations.%, limiting the variety of generated 3D structures and details. 
To address these limitations of SDS, various models have been proposed based on exploiting multi-step denoising~\cite{zhou2023sparsefusion}, a variation approach~\cite{wang2024prolificdreamer}, negative conditioning~\cite{katzir2023noise}, and ordinary differential equation trajectory~\cite{wu2024consistent3d}.
%VSD \cite{wang2024prolificdreamer} treats the 3D representation as a probabilistic distribution, optimizing multiple particles within a particle-based framework to more accurately capture the 3D distribution. 
To mitigate the limitation of noisy gradients in SDS, which hampers precise image editing, DDS \cite{hertz2023delta} was introduced. By computing the delta between the derived gradient and the target pair, DDS effectively isolates and removes unwanted noise in the gradient direction. Despite these advancements, DDS still faces challenges in preserving the complete structural consistency of the source image’s identity.
\section{Preliminaries}
\label{sec:preliminary}
\newcommand{\trg}{\mathbf{z}^{\text{trg}}}
\newcommand{\trgt}{\mathbf{z}_{t}^{\text{trg}}}
\newcommand{\src}{\mathbf{z}^{\text{src}}}
\newcommand{\srct}{\mathbf{z}_{t}^{\text{src}}}
\newcommand{\srctopt}{\mathbf{z}_{t}^{\text{src}\ast}}
\newcommand{\trgtopt}{\mathbf{z}_{t}^{\text{trg}\ast}}
\newcommand{\txts}{y^{\text{src}}} % source prompt
\newcommand{\txtt}{y^{\text{trg}}} % target prompt
\newcommand{\dds}{\text{DDS}}
\newcommand{\sds}{\text{SDS}}
\newcommand{\ids}{\text{IDS}}
\newcommand{\loss}[1]{\mathcal{L}_{#1}}
\newcommand{\grad}[1]{\nabla_{#1}}
\newcommand{\pardiff}[2]{\frac{\partial #1}{\partial #2}}
\newcommand{\postmean}{\tilde{\mathbf{z}}_{0}}

\subsection{Diffusion Model and Sampling Guidance}
Text-to-image diffusion models $\epsilon_\phi(\cdot)$ are based on diffusion probabilistic models (DPMs)~\cite{ho2020denoising, song2020score, rombach2022high}. %, which are latent variable models based on Markov chain with Gaussian distribution to synthesize the data from the distribution of the given training datasets. 
The models are trained to estimate the denoising score when the original image $\mathbf{z}_0$ and the text condition $y$ are given:
\begin{equation*}
    \mathcal{L}(\phi) = \mathbb{E}_{t, \epsilon}
    [\lVert 
    \epsilon_\phi(\mathbf{z}_t, y, t) - \epsilon 
    \rVert_2^2],
\end{equation*}
where $\epsilon\sim\mathcal{N}(0, \mathbf{I})$ and $t\sim\mathcal{U}(0, 1)$. $\mathbf{z}_t$ refers to the stochastic latent of $\mathbf{z}_0$ via the forward diffusion process as follows:
\begin{equation} \label{eq:forward}
    \mathbf{z}_t=\sqrt{\alpha_t}\mathbf{z}_0+\sqrt{1-\alpha_t}\epsilon,
\end{equation}
where $\alpha_t$ is noise schedule. With the trained $\epsilon_\phi(\cdot)$, high-quality samples can be generated using the classifier-free guidance (CFG) \cite{ho2021classifier} by subtracting unconditioned denoising score from the conditioned score with guidance scale $\omega$:
\begin{equation} \label{eq:cfg}
    \epsilon_\phi^\omega(\mathbf{z}_t, y, t) 
    = (1+\omega)\epsilon_\phi(\mathbf{z}_t, y, t) 
    -\omega\epsilon_\phi(\mathbf{z}_t, \varnothing, t).
\end{equation}

\subsection{Score Distillation Sampling (SDS)}
% \label{sec:3.2}
With pretrained text-to-image diffusion models $\epsilon_\phi(\cdot)$, SDS \cite{poole2022dreamfusion} synthesizes 3D data $\mathbf{z}$ for a given text prompt $y$ by optimizing the differentiable rendering function paremetrized by $\theta$, where $\mathbf{z}=g(\theta)$:
{\small
\begin{align}
    \grad{\theta}\loss{\sds}(\mathbf{z}, y)
    &= \mathbb{E}_{t, \epsilon} 
    \left[ \omega(t)(\epsilon_\phi^\omega(\mathbf{z}_t, y, t)-\epsilon)\pardiff{\mathbf{z}}{\theta}\right].
    \label{eq:sds}
\end{align}
}
%where $t\sim\mathcal{U}(0, 1)$ and $\epsilon\sim\mathcal{N}(0, \mathbf{I})$. 
The optimized parameters $\theta^*$ provide the text-conditioned 3D volume that follows the diffusion prior~\cite{poole2022dreamfusion}. % distribution of pretrained diffusion models. 
However, a single text prompt $y$ can refer to many different 3D volumes, each with diverse backgrounds or structural details of the object. Therefore, an inherent limitation of SDS \cite{poole2022dreamfusion} is that the score conditioned by the prompt $y$ does not always provide the diffusion prior to the identical object during the optimization process, leading to blurry and unclear results. 
%When it comes to image editing, the SDS loss in \eqref{eq:sds} is computed with respect to the data itself $\mathbf{z}=\trg$ and the target text prompt $\txtt$. In particular, $\trg$ is first set as the input image $\src$. Note that the shortcomings of SDS in image editing are even more severe, as there is no guidance regarding the source image $\src$ during the updates. As shown in Fig. \ref{fig:inversion}, the edited image $\trg$ is not only blurred, but also contains the heavily altered background and structure details that are not included in the prompt $\txtt$.

\subsection{Delta Denoising Score (DDS)}
DDS \cite{hertz2023delta} is proposed to synthesize the image $\trg$ from the given source image $\src$ and its corresponding prompt $\txts$, which is aligned to the target prompt $\txtt$. 
Based on the insight that the gradient should be zero if $\txtt$ matches $\txts$, DDS minimizes the identity change of $\src$ by simple replacing $\epsilon$ in \eqref{eq:sds} with the score $\epsilon_\phi^\omega(\srct, \txts, t)$] as follows:
{\small
\begin{equation} \label{eq:dds}
% \begin{split}
    \grad{\theta}\loss{\dds}
     = \mathbb{E}_{t, \epsilon} 
    \left[ 
      (\epsilon_\phi^\omega(\trgt, \txtt, t) - {\epsilon}_{\phi}^\omega(\srct, \txts, t)) \pardiff{\trg}{\theta}\right]. %\nonumber\\
    % = \grad{\theta}\loss{\sds}(\trg, \txtt) - \grad{\theta}\loss{\sds}(\src, \txts).
% \end{split}
\end{equation}
}
%where the same $\epsilon$ is used to generate $\trgt$ and $\srct$. 
For simplicity, we denote $\epsilon_\phi^{\text{trg}}=\epsilon_\phi^\omega(\trg, \txtt, t)$ and $\epsilon_\phi^{\text{src}}=\epsilon_\phi^\omega(\src, \txts, t)$. 
Here, $\epsilon_\phi^{\text{trg}}$ and  ${\epsilon}_\phi^{\text{src}}$ can be interpreted as the gradients representing the direction from $\trgt$ to $\trg$ and the direction from $\srct$ to $\src$, respectively.
$\theta = \trg$ is thus gradually optimized along the direction from $\src$ to $\trg$, as shown in \cref{fig:algorithm}.
It is worth noting that the guidance of the update can be calculated at the same point $\trgt$, thanks to the shared $\epsilon$.
%DDS reduces the uncertainty caused by SDS loss and provides more clear results.
However, the slight error in the gradient caused by the score ${\epsilon}_\phi^{\text{src}}$ still leads to the incorrect direction for the optimization.

\subsection{Fixed-point Iteration}

% \add{Explanation of fixed-point iteration and applications of fixed-point iteration to the diffusion model}
In numerical analysis, a fixed-point iteration \cite{parikh2014proximal} is an iterative method to find fixed points of a function $f$, where $f(x) = x.$ Given an initial point $x_0$, the iteration is defined as: 
\begin{equation*}
x_{n+1} = f(x_n), \quad n = 0, 1, 2, \ldots
\end{equation*}
Under appropriate conditions, this sequence converges to a fixed point. Thanks to its applicability to non-linear problems with low computational costs, fixed-point iteration is widely used in optimization, including applications in the context of diffusion models \cite{meiri2023fixed}. 

%These can also be applied to text-guided diffusion models, potentially enhancing their performance. Null-text Inversion \cite{mokady2023null} replaces the traditional approach of mapping all noise vectors to a single image using random noise for each optimization iteration with a more local optimization that employs a single noise vector. 
%Specifically, drawing inspiration from GAN literature, it leverages the sequence of noised latent codes obtained from an initial DDIM inversion as a pivot. The optimization is then performed around this pivot, leading to a more refined and accurate inversion. However, since this method relies on a fixed pivot, it struggles to fully preserve the complex structural consistency of the original image. We propose a novel model that utilizes fixed-point iteration to correct the gradient direction aligned with the source prompt, thereby ensuring that the structure of the source image is preserved while selectively editing the target image.
\section{Method}
\label{sec:method}

Given the source pair $\{\src, \txts \}$, the aim of our work is to provide an edited result $\trg$ that is aligned with $\txtt$ while maintaining the source's identity. To this end, we introduce a novel approach called \textbf{Identity-preserving Distillation Sampling (IDS)}, which (1) corrects the error of the gradient aligned with the text prompt by the fixed-point iterator and (2) provides the result $\trg$ using the guided noise.%$\epsilon^\ast$. %instead of random Gaussian noise for identity preservation.

%We begin with an interpretation of the SDS \cite{poole2022dreamfusion} and DDS \cite{hertz2023delta} loss function from a new perspective. 
%Although DDS \cite{hertz2023delta} has been proposed to obtain the optimal text-aligned score for image editing, structural consistency often cannot be maintained. 

%The text-conditioned score may lead to a misalignment with the identity of the given image $\src$, leading to a significant change in the overall structure and characteristics when the error is accumulated. To address this issue, we introduce a novel approach called \textbf{Identity-preserving Distillation Sampling (IDS)}, which corrects the error of the direction aligned with the text prompt by the fixed-point iterator. 

%Our method not only ensures the structural consistency of objects, but also preserves features such as color and texture. 
% when applied to image editing.

%In this section, we review and interpret DDS loss from a novel perspective in section \ref{ssec:3.1}. Based on this interpretation, we analyze the problem of what part of DDS loss is causing the error during image editing in section \ref{ssec:3.2}. Then, we introduce our method, Fixed-point Iteration for Denoising Score (FPDS), to recover this error in section \ref{ssec:3.3}.

%------------------------------------------------------------------------
% \vspace{-7mm}
\subsection{Motivation} \label{ssec:4.1}
%Therefore, the result only follows the target prompt $y_{trg}$ without preserving the characteristics of $\trg_{src}$ as shown in Fig. \ref{fig:inversion}. Is there the \textbf{score to capture the features of the input image} that should be consistent between the source and result?

%However, when editing the images, $\epsilon_\phi(\trg_t, y, t)$ generates denoising score in various directions corresponding to $y$ at each optimizing step.  

%\subsection{Problem of DDS for image editing} \label{ssec:4.2}

\textbf{Analysis of the text-conditioned score.} \ We first investigated how much identity of the given image $\src$ could be contained in the text-conditioned score ${\epsilon}_{\phi}^{\text{src}}$. To do this, we conducted the experiment to compare the original image $\src$ and the posterior mean $\src_{0|t} = \mathbb{E}[\src | \src_t]$, which is given by:
{\small
\begin{equation}
\src_{0|t} = \frac{1}{\sqrt{\alpha_{t}}}\left(
    \src_t -\sqrt{1-\alpha_{t}} {\epsilon}_{\phi}^{\text{src}}
    \right),
\label{eq:posterior}
\end{equation}
}
where $\src_t$ denotes the source latent  generated by \eqref{eq:forward}.
% As shown in the first row of Fig. \ref{fig:post_mean3}, it is difficult to recognize the features of $\src$ in $\src_{0|t}$, such as hairstyle, details of eyes, and background. This demonstrates that the score ${\epsilon}_{\phi}^{\text{src}}$ is not exactly adjusted to the given image $\src$. This deformation becomes more pronounced with increasing $t$.
As shown in the first row of the supplementary \cref{fig:post_mean3} %Fig. S1, 
it is difficult to recognize the features of $\src$ in $\src_{0|t}$, such as hairstyle, details of eyes, and background. This demonstrates that the score ${\epsilon}_{\phi}^{\text{src}}$ is not exactly adjusted to the given image $\src$. This deformation becomes more pronounced with increasing $t$.
The experiment confirms that ${\epsilon}_{\phi}^{\text{src}}$ may not be a precise guidance to the source image $\src$. %as shown in Fig. \ref{fig:algorithm}.
Therefore, the text-conditioned score ${\epsilon}_{\phi}^{\text{src}}$ needs to be modified to maintain the identity of the source image $\src$ in the edited result $\trg$.

\input{Fig./Qual/Editing/inversion}

%even if this gradient is calculated from the score $\hat{\epsilon}_{\phi}$ aligned with the prompt $\hat{y}$ that represents the given image $\src$. 
% \vspace{-10pt}
\noindent\textbf{Accumulated error in DDS.} \ The transformed image $\trg$ can be converted back to the original image $\src$ by reversing the set of $\epsilon$ used to synthesize $\trg$ from $\src$ and swapping $\{\src, \txts \}$ and $\{\trg, \txtt \}$ to calculate the DDS loss in \eqref{eq:dds}.
If the guidance from $\src$ to $\trg$ is computed exactly, the perfect reconstruction can be achieved. 
%According to our interpretation, the DDS algorithm is invertible, i.e., it is possible to reconstruct $\src$ from $\trg$, a transformed output of $\src$. 
Nevertheless, as can be seen from the second row in Fig. \ref{fig:inversion}, DDS \cite{hertz2023delta} fails to restore the original image $\src$ from the edited image $\trg$, which implies that the direction from $\src$ to $\trg$ is calculated incorrectly.
Based on our analysis, this error is because the text-conditioned score ${\epsilon}_\phi^{\text{src}}$ do not refer to the source $\src$, which can be explictly expressed as the difference between the injected noise $\epsilon$ and the score ${\epsilon}_\phi^{\text{src}}$. While the optimization is being processed, the error inevitably accumulates, leading to the undesirable change to the structure and the pose. 
% To address these issues, we investigated whether the guidance from $\src$ to $\trg$ can be properly provided while preserving the source's identity, when the timestep $t$ is constrained by $t\sim\mathcal{U}(0, 0.2)$. This is because the posterior mean $\src_{0|t}$ and the source image $\src$ are similar for small timestep $t$, as illustrated in the first row of Fig. \ref{fig:post_mean3}. However, as depicted in the first row of Fig. \ref{fig:inversion}, DDS yields unrealistic result with this setting, whereby the structure of the given image $\src$ is overemphasized. This implies that it is not sufficient to simply limit the timestep $t$ to prevent the score from deviating too far from $\src$ to correct the misalignment of the score to $\src$.
To address these issues, we investigated whether the guidance from $\src$ to $\trg$ can be properly provided while preserving the source's identity, when the timestep $t$ is constrained by $t\sim\mathcal{U}(0, 0.2)$. This is because the posterior mean $\src_{0|t}$ and the source image $\src$ are similar for small timestep $t$, as illustrated in the first row of supplementary \cref{fig:post_mean3}. %Fig. S1. 
However, as depicted in the first row of Fig. \ref{fig:inversion}, DDS yields unrealistic result with this setting, whereby the structure of the given image $\src$ is overemphasized. This implies that it is not sufficient to simply limit the timestep $t$ to prevent the score from deviating too far from $\src$ to correct the misalignment of the score to $\src$.
Hence, we propose a fundamental approach to refine the gradient to achieve identity consistency without unwanted overemphasis on details.

%To accurately calculate the direction from $\src$ to $\trg$, the direction of estimated denoising score $\grad{\theta}\hat{\epsilon}_\phi$ should be equal to the direction from $\src_t$ to $\src$ that is determined by $\epsilon$. 
%In DDS, however, $\hat{\epsilon}_\phi$ is not equal to $\epsilon$ since $\epsilon$ is randomly sampled value, and $\hat{\epsilon}_\phi$ is denoising score when $\src_t$ corresponds to the prompt $\hat{y}$. The difference between $\hat{\epsilon}_\phi$ and $\epsilon$ makes misalignment for the direction of $\grad{\theta}\loss{\dds}$, and this error accumulates during the optimization process. Although DDS shows meaningful performance for image-to-image translation, this cumulated error causes DDS to edit the background or structure incorrectly. %manifold fig?
% $\trg$ is the result edited from source image $\src$ and prompt $\hat{y}$ to target prompt $y$, and $\src^{\dagger}$ is the inverted result that transforms the edited image $\trg$ for prompt $y$ to previous source prompt $\hat{y}$. 
% However, as shown in Fig. \ref{fig:inversion}, the inversion $\src^{\dagger}$ of DDS is not same as the given source image due to the misalignment between $\hat{\epsilon}_\phi$ and $\epsilon$ which is used to make the input of $\epsilon_\phi$.

% In Fig. \ref{fig:inversion}, 
% That is, the reverse of \ref{eq:forward} should be similar with $\src$:
% \begin{equation}
%     \postmean \approx \frac{1}{\sqrt{\alpha_t}}(\src_t - \sqrt{1-\alpha_t} \hat{\epsilon}_\phi)
% \end{equation}

\subsection{Identity-preserving Distillation Sampling (IDS)} \label{ssec:4.2}

\textbf{Fixed-point Regularization (FPR).} \
Here, we introduce a \textbf{F}ixed-\textbf{p}oint \textbf{R}egularization (FPR) method that adjusts the text-conditioned score ${\epsilon}_\phi^{\text{src}}$ to the source image $\src$. Our key premise is that if the score ${\epsilon}_\phi^{\text{src}}$ is rightly estimated as a gradient to $\src$, the posterior mean $\src_{0|t}$ also contains sufficient information about $\src$. %, resulting in identity preservation.
Therefore, FPR loss is designed to minimize the difference between $\src$ and $\src_{0|t}$ as follows:
\begin{equation}
\mathcal{L}_{\text{FPR}} = d ( \mathbf{z}^{\text{src}}, \mathbf{z}_{0|t}^\text{src}),
\label{eq:fpr}
\end{equation}
%\begin{equation}
%    \mathcal{L}_{\text{FPR}}
%    =\lVert \mathbf{z}_{0|t}^\text{src} - \mathbf{z}^{\text{src}} \rVert _{2}^{2},  
%\end{equation}
where $d(\mathbf{x}_1, \mathbf{x}_2)$ can be any metric to compare $\mathbf{x}_1$ and $\mathbf{x}_2$. Here, we employed the Euclidean loss, and further investigations using various metrics are provided in \cref{sec:s_metricsforfpr} of Supplementary Materials.

The score ${\epsilon}_\phi^{\text{src}}$ needs to be modified to minimize the FPR loss before obtaining the updated direction. There are two ways to control the score  ${\epsilon}_\phi^{\text{src}}$ by altering the injection noise $\epsilon$ or the source latent $\mathbf{z}^{\text{src}}_t$. 
% As illustrated in Fig.~\ref{fig:post_mean3}, the proposed FPR revises the score ${\epsilon}_\phi^{\text{src}}$ to serve the source's identity for both approaches. Note that the score incorporates the content details, with the updates being performed with respect to the source latent $\mathbf{z}^{\text{src}}_t$ compared to the noise $\epsilon$. Thus, $\mathbf{z}_t^{\text{src}}$ is updated to minimize the FPR loss as follows:
As illustrated in supplementary \cref{fig:post_mean3}, %Fig. S1, 
the proposed FPR revises the score ${\epsilon}_\phi^{\text{src}}$ to serve the source's identity for both approaches. Note that the score incorporates the content details, with the updates being performed with respect to the source latent $\mathbf{z}^{\text{src}}_t$ compared to the noise $\epsilon$. Thus, $\mathbf{z}_t^{\text{src}}$ is updated to minimize the FPR loss as follows:
\begin{equation}\label{eq:update}
    \mathbf{z}^{\text{src}}_t \leftarrow \mathbf{z}_{t}^{\text{src}} - \lambda \nabla_{\mathbf{z}^{\text{src}}_t}\mathcal{L}_{\text{FPR}},
\end{equation}
where $\lambda$ and $N$ denote a regularization scale and the number of iterations, respectively. 

\newcommand{\bfz}{\mathbf{z}}
\newcommand{\hbfz}{\hat{\mathbf{z}}}
\begin{algorithm}
\caption{Fixed-point Regularization (FPR)}
\label{alg:fpr}
\begin{algorithmic}[1]
%\REQUIRE source image $\mathbf z^{\text{src}}$, source prompt ${y}^{\text{src}}$, timestep $t$, diffusion model $\epsilon_{\phi}$, scale (CFG)  $w$, scale (IDS) $a$
%\ENSURE Output $\epsilon^{\ast}$ %target image $\bfz$

\REQUIRE $\src$, $\txts$, $\epsilon_{\phi}$, $\omega$, $\lambda$, $N$
%\ENSURE Output $\epsilon^{\ast}$ %target image $\bfz$

\STATE $\epsilon \sim \mathcal{N}(0, \mathbf{I})$
\STATE $t \sim\mathcal{U}(0, 1)$
\STATE $\mathbf z^{\text{src}}_t \leftarrow \sqrt{\alpha_t}\mathbf z^{\text{src}}+\sqrt{1-\alpha_t}\epsilon$
\FOR{i = 1, $\dots$, N}
    \STATE $\epsilon_\phi^\text{src} \leftarrow 
    (1+\omega)\epsilon_\phi(\mathbf z^{\text{src}}_{t}, y^{\text{src}}, t)
    - \omega \epsilon_\phi(\mathbf z^{\text{src}}_t, \varnothing, t)$
    \STATE $\bfz_{0|t}^{\text{src}} \leftarrow \frac{1}{\sqrt{\alpha_t}}
    (\mathbf z^{\text{src}}_t - \sqrt{1-\alpha_t}\epsilon_\phi^\text{src})$
    \STATE $\mathcal{L}_{\text{FPR}} \leftarrow 
     d(\src_{0|t}, \src)$
    \STATE $\mathbf z^{\text{src}}_t \leftarrow \mathbf z^{\text{src}}_t - \lambda \nabla_{\mathbf z^{\text{src}}_t}\mathcal{L}_{\text{FPR}}$
\ENDFOR

\STATE $\epsilon^\ast \leftarrow 
\frac{1}{\sqrt{1-\alpha_t}}(\mathbf z^{\text{src}}_t - \sqrt{\alpha_t}\mathbf z^{\text{src}})$
% \STATE $\bfz \leftarrow \text{DDS}(\hbfz, \hat{y}, y, \epsilon^{\ast})$
\RETURN $\epsilon^\ast$
\end{algorithmic}
\end{algorithm}
% \vspace{-6pt}
%Figure 5.1
\input{Fig./Qual/Editing/figure_ip2p_sub}
% \vspace{-10pt}
\noindent\textbf{Editing with guided noise.} \ 
Thanks to the proposed FPR, the optimized source latent $\srctopt$ containing the source's identity can be obtained. 
Then, the guided noise $\epsilon^\ast$ is extracted as follows:
{\small
\begin{equation} \label{eq:guide_eps}
\epsilon^\ast = \frac{1}{\sqrt{1-\alpha_t}}(\srctopt - \sqrt{\alpha_t} \src).
\end{equation}
} 
$\epsilon^\ast$ is utilized to produce the stochastic latent $\trgtopt$ by applying the forward diffusion process to the target image $\trg$.
With $\srctopt$ and $\trgtopt$, the updated direction is given by:
{\small
\begin{equation} \label{eq:ids}
% \begin{split}
    \grad{\theta}\loss{\ids}
     = \mathbb{E}_{t, \epsilon} 
    \left[ 
      (\epsilon_\phi^\omega(\trgtopt, \txtt, t) - {\epsilon}_{\phi}^\omega(\srctopt, \txts, t)) \pardiff{\trg}{\theta}\right]. %\nonumber\\
    % = \grad{\theta}\loss{\sds}(\trg, \txtt) - \grad{\theta}\loss{\sds}(\src, \txts).
% \end{split}
\end{equation}
}
It is worth noting that $\epsilon^\ast$ guides the appropriate gradients for editing while conserving the source's identity. In contrast to DDS, the proposed IDS perfectly reconstructs the source from the edited result $\trg$, as shown in the third row of Fig.~\ref{fig:inversion}. 
This confirms that the correct score and the corresponding injection noise can preserve the identity without further consideration of mutual information. The flowchart of our IDS is illustrated in Fig.~\ref{fig:teasor}.

\section{Results} \label{sec:results}
We evaluate our method through editing experiments conducted on two experiments. In \cref{sec:5.1}, we perform a comparison on image-to-image editing across several datasets. In \cref{sec:5.2}, we extend our evaluation to editable Neural Radiance Fields (NeRF) \cite{mildenhall2021nerf}.

\subsection{Text-guided image editing}
\label{sec:5.1}
\noindent\textbf{Baselines.} To evaluate our method, we conduct comparative experiments against four state-of-the-art image editing models: Prompt-to-Prompt (P2P) \cite{hertzprompt}, Plug-and-Play (PNP) \cite{tumanyan2023plug}, DDS \cite{hertz2023delta}, and CDS \cite{nam2024contrastive}. The implementations of the baselines are carried out by referencing the official source code for each method. More details are provided in \cref{sec:s_implement} of Supplementary Materials.

\noindent\textbf{Qualitative Results.} We present the qualitative results comparing our method with the baselines in \cref{fig:ip2p_qual}. Prompt-to-Prompt (P2P) \cite{hertzprompt} performs image editing after applying DDIM inversion \cite{dhariwal2021diffusion, song2020denoising} to the source image, leading to disregarding the structural components of the source image and following the target prompt excessively. Plug-and-Play (PnP) \cite{tumanyan2023plug} has limitations in object recognition, as seen in the fourth row of Fig.~\ref{fig:ip2p_qual}. The third row of Fig.~\ref{fig:ip2p_qual} demonstrates that DDS \cite{hertz2023delta} and CDS \cite{nam2024contrastive} exhibited limitations, particularly in preserving the structural characteristics of the source image. In contrast, our method successfully edits the image while preserving the structural integrity of the source image.
% exhibit limitations such as failing to maintain the handle length and saddle shape of the bike in the first row and being unable to preserve the structure of the shark in the second row. %Furthermore, as seen in the third and fourth rows, the details in the edited target areas lacked refinement, and in the last row, the color of the source image was not preserved. In contrast, our method successfully edits the image aligning with the target text prompt while preserving the structural integrity of the source image.

\noindent\textbf{Quantitative Results.} 
% We employed two datasets: LAION 5B \cite{schuhmann2022laion} and InstructPix2Pix \cite{brooks2023instructpix2pix}.
% ##ORIGINAL## To measure the identity-preserving performance, we utilize two datasets. First, we collect 250 cat images from the LAION 5B dataset \cite{schuhmann2022laion} based on \cite{nam2024contrastive} for \textit{Cat-to-Others} task. We measure Intersection over Union (IoU) to evaluate how much of the area of the source object has been preserved. Second, we gather 28 images from the InstructPix2Pix (IP2P) dataset \cite{brooks2023instructpix2pix}, which contains the pairs of source and target images and corresponding prompts. We calculate the background Peak-Signal-to-Noise-Ratio (PSNR) to assess how the identity of the source image is preserved after editing. In addition, we use the LPIPS score \cite{zhang2018unreasonable} for each experiment to quantify the similarity between source and target images. The results are presented in \cref{tab:2Dquan}. Our method consistently achieves the lowest LPIPS score across all datasets, indicating that it best preserves the structural semantics of the source images. 
To measure the identity-preserving performance, we utilize two datasets. First, we collect 250 cat images from the LAION 5B dataset \cite{schuhmann2022laion} based on \cite{nam2024contrastive} for \textit{Cat-to-Others} task and measure Intersection over Union (IoU). Second, we gather 28 images from the InstructPix2Pix (IP2P) dataset \cite{brooks2023instructpix2pix}, which contains the pairs of source and target images and corresponding prompts and calculate the background Peak-Signal-to-Noise-Ratio (PSNR). Details of the metrics are provided in Supplementary Materials \cref{sec:s_evalmetric}. In addition, we use the LPIPS score \cite{zhang2018unreasonable} for each experiment to quantify the similarity between source and target images. The results are presented in \cref{tab:2Dquan}. Our method consistently achieves the lowest LPIPS score across all datasets, indicating that it best preserves the structural semantics of the source images. 
% We collect 250 images of cats from the LAION 5B dataset \cite{schuhmann2022laion} based on \cite{nam2024contrastive} for \textit{Cat-to-Others} task and 28 images from the InstructPix2Pix dataset \cite{brooks2023instructpix2pix} following the regulations. To evaluate the images translated by each method, we measure Intersection over Union (IoU) on LAION 5B, which primarily consists of object-focused data. We also measure the background PSNR on InstructPix2Pix to assess the extent to which the source image’s identity is preserved after editing. The results are presented in \cref{tab:2Dquan}. 
% Our method consistently achieves the lowest LPIPS score across all datasets, indicating that it best preserves the structural semantics of the source images. 
\begin{table}[b]
\centering
\resizebox{0.98\columnwidth}{!}{
\small{
\begin{tabular}{c|cc|cc|cc}
\hline
& \multicolumn{2}{c|}{cat2pig} & \multicolumn{2}{c|}{cat2squirrel} & \multicolumn{2}{c}{Ip2p}  \\ 
\hline
\multicolumn{1}{c|}{Metric} & IoU ($\uparrow$) & LPIPS ($\downarrow$) & IoU ($\uparrow$) & LPIPS ($\downarrow$) & PSNR ($\uparrow$) & LPIPS ($\downarrow$) \\ 
\hline
P2P \cite{hertzprompt}& 0.58 & 0.42 & 0.52 & 0.46 & 20.88 & 0.47 \\
PnP \cite{tumanyan2023plug}& 0.55 & 0.52 & 0.53 & 0.52 & 23.81 & 0.39 \\
DDS \cite{hertz2023delta}& 0.69 & 0.28 & 0.65 & 0.30 & 26.02 & 0.24 \\  
CDS \cite{nam2024contrastive}& 0.72 & 0.25 & \textbf{0.71} & 0.26 & 27.35 & 0.21 \\
\hline
\textbf{IDS (Ours)} & \textbf{0.74} & \textbf{0.22} & \textbf{0.71} & \textbf{0.24} & \textbf{29.25} & \textbf{0.19} \\
\hline
\end{tabular}
}
}
\vspace{-5pt}
\caption{\textbf{Quantitative results} for image editing. LPIPS \cite{zhang2018unreasonable} and IoU was measured on LAION 5B \cite{schuhmann2022laion}, while LPIPS and background PSNR was measured on InstructPix2Pix \cite{brooks2023instructpix2pix}.}
\label{tab:2Dquan}
\end{table}

%P2P \cite{hertzprompt}& 0.5798 & 0.4229 & 0.5184 & 0.4605 & 20.88 & 0.4695 \\
%PnP \cite{tumanyan2023plug}& 0.5507 & 0.5191 & ??? & 0.5245 & 23.81 & 0.3882 \\
%DDS \cite{hertz2023delta}& 0.6897 & 0.2838 & 0.6456 & 0.2996 & 26.02 & 0.2398 \\  
%CDS \cite{nam2024contrastive}& 0.7249 & 0.2485 & 0.7054 & 0.2612 & 27.35 & 0.2099 \\

\begin{table}[bh!]
\vspace{-5pt}
\centering
%\scalebox{0.65}
\resizebox{1.0\columnwidth}{!}{
%\small{ %
\begin{tabular}{c|ccc|ccc}
\hline
& \multicolumn{3}{c|}{User Preference Rate (\%)} & \multicolumn{3}{c}{GPT score \cite{peng2024dreambench++}}\\ 
\hline
\multicolumn{1}{c|}{Metric} & Text ($\uparrow$) & Preserving ($\uparrow$) & Quality ($\uparrow$) & Text ($\uparrow$) & Preserving ($\uparrow$) & Quality ($\uparrow$) \\ 
\hline
P2P \cite{hertzprompt}& 11.13 & 4.80 & 8.09 & 5.66 & 5.37 & 5.77 \\
PnP \cite{tumanyan2023plug}& 7.72 & 7.17 & 6.93 & 6.54 & 6.77 & 6.74 \\
DDS \cite{hertz2023delta}& 20.30 & 10.82 & 16.23 & 7.60 & 7.51 & 7.37 \\
CDS \cite{nam2024contrastive}& 17.02 & 16.72 & 17.08 & 8.26 & 8.00 & 8.09 \\ 
\hline
\textbf{IDS (Ours)} & \textbf{43.83} & \textbf{60.49} & \textbf{51.67} & \textbf{8.97} & \textbf{9.00} & \textbf{8.80} \\
\hline
\end{tabular}
}
%}
\vspace{-5pt}
\caption{\textbf{User study and GPT scores}  \cite{peng2024dreambench++} show that our method achieved the highest scores across all questions for image editing.}
\label{tab:Userstudy_GPTscore}
\end{table}
For user evaluation, we present 35 comparison sets for four baselines and our method, gathering responses from 47 participants. Participants are asked to choose the most appropriate image for the following three questions: 1. \textit{Which image best fits the text condition?} 2. \textit{Which image best preserves the structural information of the original image?} 3. \textit{Which image has the best quality for text-based image editing?} 
Additionally, we measure the GPT score using the Dreambench++ \cite{peng2024dreambench++} method, which generates human-aligned assessments for the same questions by refining the scoring into ten distinct levels. As shown in \cref{tab:Userstudy_GPTscore}, our method receives the highest ratings for all questions.
% Furthermore, we ask users to select their favorite image from the baselines in order to gauge their preferences, and we compute the selected ratio in percentage terms.
%While our CLIP score was not significantly higher than other methods, it remained comparable. %Considering the outcomes of both metrics, our model demonstrates an ability to maximally preserve the source image's structure during the editing process while minimally and precisely transforming the regions specified by the target prompt.

% Fig 5.2

\input{Fig./Qual/Editing/figure_FICUS}
\subsection{Editing NeRF}
We conduct experiments involving 3D rendering of edited images to demonstrate the effectiveness of our method in maintaining structural consistency. This approach is particularly relevant as consistency has an even greater impact on outcomes in 3D environments.

\label{sec:5.2}

\noindent\textbf{Datasets.} We evaluated our method on widely used NeRF datasets: Synthetic NeRF \cite{mildenhall2021nerf} and LLFF \cite{mildenhall2019local}. Since NeRF datasets have no given pairs of source and target prompts, we manually composed image descriptions.
%, such as the source prompt ``A tree in a brown vase" and its corresponding target prompt ``A tree in a blue vase" as shown in \cref{fig:ficus_qual}.

\noindent\textbf{Qualitative Results.} \cref{fig:ficus_qual} illustrates the qualitative results of our method compared with NeRF editing baselines. In the first row, the target prompt specifies a precise part of the image for fine-grained editing. DDS \cite{hertz2023delta} and CDS \cite{nam2024contrastive} fail to differentiate and edit the specific area. At the same time, our method accurately identifies the region indicated by the target prompt in the image and performs detailed editing exclusively on that part. 
The second row demonstrates a scenario in which the target prompt is designed to edit the mood of the image. Our approach adjusts the colors associated with ``autumn" and ``leaves" throughout the image while maintaining consistency in the ``trunk" whereas DDS and CDS also changed the ``trunk". In terms of depth maps, our method generates clean depth maps with minimal noise after image editing, whereas DDS and CDS introduce noticeable noise into the depth maps.

\begin{table}[thb!]
\centering
\resizebox{0.95\columnwidth}{!}{
\begin{tabular}{c|c|ccc}
\hline
\multirow{2}{*}{Metric} & \multirow{2}{*}{CLIP \cite{radford2021learning}  ($\uparrow$)} & \multicolumn{3}{c}{User Preference Rate (\%)} \\ 
\cline{3-5}
& & Text ($\uparrow$) & Preserving ($\uparrow$) & Quality ($\uparrow$) \\ 
\hline
DDS \cite{hertz2023delta}& 0.1596 & 36.88 & 28.37 & 32.62 \\
CDS \cite{nam2024contrastive}& 0.1597 & 22.70 & 23.40 & 21.28 \\
\hline
\textbf{IDS (Ours)} & \textbf{0.1626} & \textbf{40.42} & \textbf{48.23} & \textbf{46.10} \\
\hline
\end{tabular}
}
\caption{\textbf{Quantitative results of NeRF editing} with respect to CLIP score and User Preference Rate. IDS demonstrates superior quantitative performance compared to the baselines.}
\label{tab:Nerfclip}
\end{table}

\noindent\textbf{Quantitative Results.} Based on edited images, we performed 3D rendering and subsequently conducted quantitative evaluations provided in \cref{tab:Nerfclip}. To assess whether the edited 3D images are precisely aligned with the target prompts, we measured the CLIP \cite{radford2021learning} scores at 200k iterations of training on the LLFF dataset. We additionally present a user evaluation conducted under the same setup in \cref{sec:5.1}. Consistent with the trends observed in the qualitative results, our method demonstrates superior performance in the quantitative evaluations compared to other baselines.
%To demonstrate the effectiveness of our method in maintaining structural consistency during image editing and correcting errors progressively throughout training, we also conduct experiments involving 3D rendering of edited images. This approach is particularly relevant as consistency has an even greater impact on outcomes in 3D environments.

% \vspace{-10pt}
% \input{Fig./Qual/Editing/fp_iter}
% \vspace{-10pt}
\input{Fig./Qual/Editing/scale}
% \vspace{-5pt}
\input{Fig./Qual/Editing/400steps}
\begin{table}[t]
\centering
\resizebox{0.98\columnwidth}{!}{
\small{
\begin{tabular}{c|cc|cc|c|c}
\hline
& optim step & FPR iter & LPIPS($\downarrow$) & CLIP($\uparrow)$ & time (sec/img) & Memory (GB)  \\ 
\hline
% DDS & \multirow{2}{*}{200} & \multirow{2}{*}{-} & 0.240 & 0.293 & 22.45 & 6.27 \\
%CDS &                      &                    & 0.210 & 0.287 & 59.31 & 8.83 \\
DDS & 200 & - & 0.240 & 0.293 & 22.45 & 6.27 \\
\hline
CDS &  200 &  -  & 0.210 & 0.287 & 59.31 & 8.83 \\
\hline
\multirow{4}{*}{IDS} & \multirow{2}{*}{200} & 1 & 0.199 & 0.285 & 50.80 & \multirow{4}{*}{8.63} \\
                     &                      & 3 & 0.190 & 0.277 & 107.77 & \\
\cline{2-6}
& 100 & \multirow{2}{*}{3} & 0.165 & 0.265 & 54.04 & \\
& 150 &                    & 0.180 & 0.272 & 81.25 & \\
\hline
\end{tabular}
}
}
\vspace{-5pt}
\caption{\textbf{Computational complexity} on 28 images of InstructPix2Pix \cite{brooks2023instructpix2pix} for various settings. Lower LPIPS and higher CLIP scores mean better quality.}
% \vspace{-3pt}
\label{tab:overhead}
\end{table}
\section{Discussions}
% \subsection{Harmony with Existing SDS}
% IDS can be applied to SDS optimization for a given source image and prompt to preserve the original contents and reduce the blurry effect. As shown in \cref{fig:existingSDS}, the conserved rate of the information of the source image is controllable by the number of iterations of FPR.
% when using IDS to SDS optimization, the original structural information of the source image is maintained by fixing the imperfect score caused by insufficient prompt and random noise $\epsilon$. 
% \vspace{-15pt}
% \input{Fig./Qual/Editing/existingSDS}
% We conduct ablation studies to demonstrate the effectiveness of different fine-tuning strategies within our method.
% \input{Fig./Qual/Editing/400steps}
% \vspace{-15pt}
% \input{Fig./Qual/Editing/scale}

% 6.1 images
\subsection{Ablation studies on FPR}
\noindent\textbf{FPR iteration $N$.} 
% In the proposed FPR, the number of iterations $n$ is one of the hyper-parameters.
%In our method, Fixed-point iteration refers to the number of times the fixed-point process is applied when correcting $\epsilon$ errors. 
We conduct experiments on FPR iterations $N$ to evaluate its impact and determine the optimal iteration count. Although performing just one iteration of FPR is sufficient to preserve the source identity, as shown in lower LPIPS score than baselines of \cref{tab:overhead}, we set $N=3$ to emphasize the purpose of our method.
%We set $n=3$ based on the trade-off between the computational overhead and preserving identity.
%However, for finer details, such as editing the shape of a ``pig nose", a higher iteration count of three or more yields the best results. Considering both execution time and efficiency, we set the Fixed-point iteration count to three in our method.
% \vspace{-11pt}
%This ensures that even with an increased number of inference steps, the structural integrity of the source image is maintained, enabling precise image editing aligned with the target prompt.
% \vspace{-11pt}

\noindent\textbf{Scale $\lambda$.}
The scaling factor $\lambda$ of FPR determines how much information of source latent $\src$ is kept. As shown in \cref{fig:ablation_scale}, increasing the scale preserves the attributes of the source image, resulting in more successful editing when it is hard to translate due to the structural mismatch between the source and the target prompt.
% The scaling factor $\lambda$ of the FPR determines how much the source latent $\srct$ is modified depending on the $d(\src, \src_{0|t})$.
%when calculating gradients based on the posterior mean to correct $\epsilon$ errors. 
% As shown in \cref{fig:ablation_scale}, increasing the scale enhances both the structural and color fidelity of the source image, highlighting the importance of the scale factor in preserving the source image’s structural attributes. %Thus, the scale $\lambda$ is set to 1 in all experiments.
% \vspace{-13pt}

% \vspace{-10pt}
%\paragraph{Optimization steps.}
\subsection{Optimization steps}
To show that our method can prevent error accumulation during translation, we set the experiment to extend the number of optimization steps from 200 to 400. In the results of DDS and CDS, there is color boosting or loss of details due to the cumulated error. In contrast, IDS maintains the characteristics of the original images, such as the color of the pumpkin in the first row of \cref{fig:ablation_step} and the shape of the leaf in the second row of \cref{fig:ablation_step}, better than other methods.

\section{Limitation}
\label{sec:limit}
\begin{figure}[H] % 1-column
\centering
\footnotesize

% Adjust the image width so that 5 images fit within one column
\newcommand{\imgwidth}{0.235\linewidth}

% 1st Image
\begin{tikzpicture}[x=1cm, y=1cm]
    \node[anchor=south] (FigA1) at (0,0) {
        \includegraphics[width=\imgwidth]{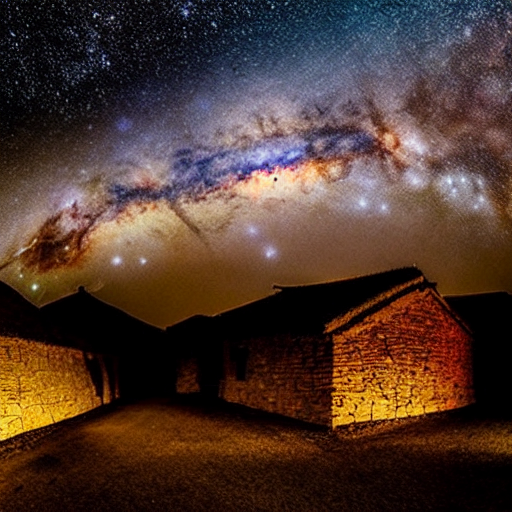}
    };
    \node[anchor=south, yshift=-1mm] at (FigA1.north) {\footnotesize Source};
\end{tikzpicture}\hspace{-1mm}%
% 2nd Image
\begin{tikzpicture}[x=1cm, y=1cm]
    \node[anchor=south] (FigB1) at (0,0) {
        \includegraphics[width=\imgwidth]{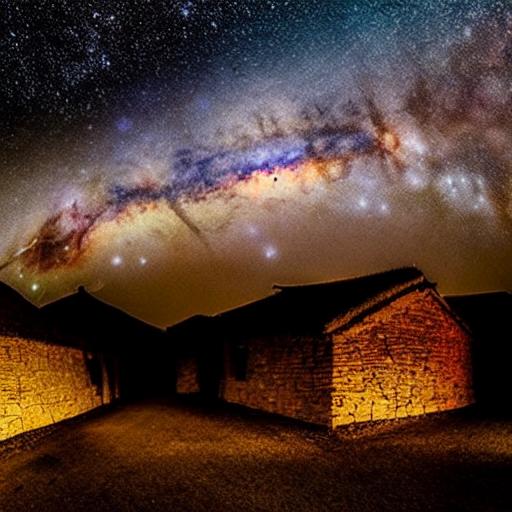}
    };
    \node[anchor=south, yshift=-1mm] at (FigB1.north) {\footnotesize \textbf{IDS (Ours)}};
\end{tikzpicture}\hspace{-1mm}%
% 3rd Image
\begin{tikzpicture}[x=1cm, y=1cm]
    \node[anchor=south] (FigC1) at (0,0) {
        \includegraphics[width=\imgwidth]{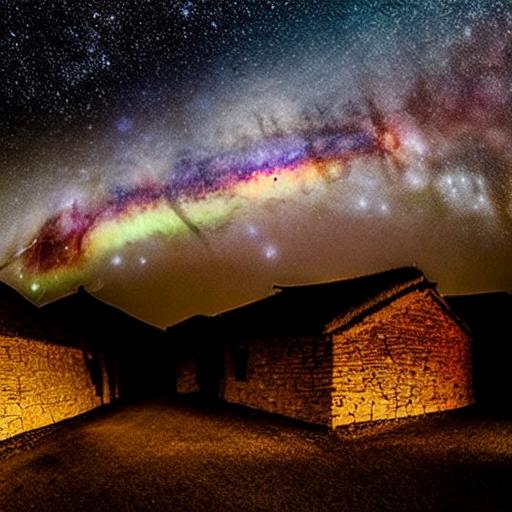}
    };
    \node[anchor=south, yshift=-1mm] at (FigC1.north) {\footnotesize CDS};
\end{tikzpicture}\hspace{-1mm}%
% 4th Image
\begin{tikzpicture}[x=1cm, y=1cm]
    \node[anchor=south] (FigD1) at (0,0) {
        \includegraphics[width=\imgwidth]{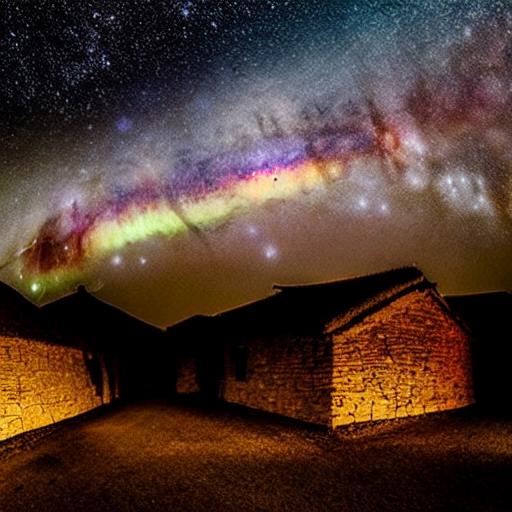}
    };
    \node[anchor=south, yshift=-1mm] at (FigD1.north) {\footnotesize DDS};
\end{tikzpicture}\hspace{-1mm}%

\vspace{-3pt}
\setulcolor{magenta}
\setul{0.3pt}{2pt}
\centering
\textit{``Photo free night, house, aurora" $\to$ ``... \ul{with two dogs}"} 
\vspace{-10pt}
\caption{\textbf{Failure case} for complex text prompt.}
\vspace{-8pt}
\label{fig:complex_ex}
\end{figure}

%The proposed IDS effectively preserves source's identity, including pose, structure and background, by iteratively updating the source latent $\srct$ according to the given source information. As represented in \ref{sec:results}, 
% The proposed IDS demonstrates outstanding performance across evaluation metrics assessing consistency between source and target images. However, during FPR process, IDS relies solely on information from the source ($\{ \src, y^{\text{src}} \}$) without incorporating target-side information. This results in comparatively lower CLIP scores~\cite{radford2021learning} than other baselines as reported in Tab.~\ref{tab:limitation} and failure cases for more complex translations as shown in. In addition, our method requires additional computational overhead since FPR is applied to each optimization iteration. Detailed discussion about our limitation is provided in Supp.
The proposed IDS demonstrates outstanding performance across evaluation metrics assessing consistency between source and target images. However, during FPR process, IDS relies solely on information from the source ($\{ \src, y^{\text{src}} \}$) without incorporating target-side information. This results in comparatively lower CLIP scores~\cite{radford2021learning} than other baselines (\cref{tab:limitation}) and failure cases for more complex translations (\cref{fig:complex_ex}). In addition, our method requires additional computational overhead (\cref{tab:overhead}) since FPR is applied to each optimization iteration. Detailed discussion about our limitation is provided in \cref{sec:supp_limit} of supplementary.
Our future direction will explore changing the score conditioned by the target prompt $y^{\text{trg}}$, leading to a better alignment with $y^{\text{trg}}$.
% which is iteratively obtained by \cref{eq:update}. The equation only uses information from a source (prompt \& image), not from a target side. This fact hampers text-driven representations of a target prompt leading to lower CLIP scores \cite{radford2021learning} compared with other baselines.
\begin{table}[thb!]
\centering
\resizebox{0.95 \columnwidth}{!}{
\normalsize{
\begin{tabular}{c|c c c c | c}
\hline
&  P2P \cite{hertzprompt} & PnP \cite{tumanyan2023plug} & DDS \cite{hertz2023delta} & CDS \cite{nam2024contrastive} & \textbf{IDS (Ours)} \\
\hline
\textbf{cat2lion} & 0.29 & 0.21 & \textbf{0.30} & 0.29 & 0.29 \\
\textbf{cat2dog} & \textbf{0.27} & 0.26 & \textbf{0.27} & \textbf{0.27} & 0.26\\
\textbf{Ip2p}  & 0.28 & \textbf{0.30} & 0.29 & 0.29 & 0.28\\
\hline
\end{tabular}
}
}
\vspace{-5pt}
\caption{\textbf{Limitation of IDS with respect to CLIP score\cite{radford2021learning}} for image editing on LAION 5B \cite{schuhmann2022laion} and InstructPix2Pix \cite{brooks2023instructpix2pix}.} %A higher value means better performance.}
\vspace{-20pt}
\label{tab:limitation}
\end{table}

%P2P \cite{hertzprompt}& 0.2941 & \textbf{0.2689} & 0.2797 \\
%PnP \cite{tumanyan2023plug}& 0.2096 & 0.2621 & \textbf{0.3038} \\
%DDS \cite{hertz2023delta}& \textbf{0.2972} & 0.2677 & 0.2930 \\  
%CDS \cite{nam2024contrastive}& 0.2940 & 0.2667 & 0.2871 \\
%\hline
%\textbf{IDS (Ours)} & 0.2870 & 0.2629 & 0.2774 \\

\section{Conclusion}
We proposed a new distillation sampling method using a fixed-point regularization which aligns the text-conditioned score towards identity-preserved manifolds. The proposed fixed-point regularization preserves the source's identity by re-projecting the intermediate score status on posterior means. In this manner, corrected noises guide a gradient of distilled score toward identity-consistent manifolds. Owing to self-correction by a fixed-point iterator and guided injection noise, the proposed identity-preserving distillation sampling provides clear and unambiguous representations corresponding to the given prompts in text-guided image editing and editable neural radiance field (NeRF). Furthermore, our model can be utilized as a universal module in addition to the existing score-sampling processes. %Inverted initial noise from pre-trained diffusion model is clearly retrieved by our method.
\newpage

\paragraph{Acknowledgement} This work was supported by the National Research Foundation of Korea(NRF) grant funded by the Korea government(MSIT) (RS-2024-00335741, RS-2024-00357197).

{
    \small
    \bibliographystyle{ieeenat_fullname}
    \bibliography{main}
}

\clearpage
\setcounter{page}{1}
\maketitlesupplementary

\setcounter{section}{0}
\renewcommand\thesection{\Alph {section}}

\renewcommand{\thefigure}{S\arabic{figure}}
\setcounter{figure}{0}

\renewcommand{\thetable}{S\arabic{table}}
\setcounter{table}{0}

\section{Posterior mean analysis}
\label{sec:s_postmean}

To investigate how much identity of the original image $\mathbf{z}$ is contained in the text-conditioned score $\epsilon_{\phi}(\mathbf{z}, y, t)$, we conduct the experiment in which the posterior mean is obtained from various timesteps. As shown in the first row of \cref{fig:post_mean3}, more primary information is damaged as the timestep $t$ increases. On the other hand, when using FPR, since the score $\epsilon_\phi(\mathbf{z}, y, t)$ is modified to preserve the identity of $\mathbf{z}$, we can see that it has more information than before, even at large timestep, as described in the second and third row of \cref{fig:post_mean3}. Note that the score $\epsilon_\phi(\mathbf{z}, y, t)$ can be controlled by updating the injection noise $\epsilon$ or the noisy latent $\mathbf{z}_t$. Of the two options, it has been updated for $\mathbf{z}_t$ because it contains more content details.

\input{Fig./Qual/Editing/post_mean3}

\section{Metrics for FPR}
\label{sec:s_metricsforfpr}

\begin{figure}[thb!] % 1-column
\footnotesize
\centering 

% Adjust the image width to fit within one column
\newcommand{\imgwidth}{0.235\linewidth} % Set image width to 16% of the line width

% \hspace{-1.9mm}
% \raisebox{0.25in}{\rotatebox{90}{Source}}%
% \hspace{-0.9mm}
% 1st Image
\begin{tikzpicture}[x=1cm, y=1cm]
    \node[anchor=south] (FigA1) at (0,0) {
        \includegraphics[width=\imgwidth]{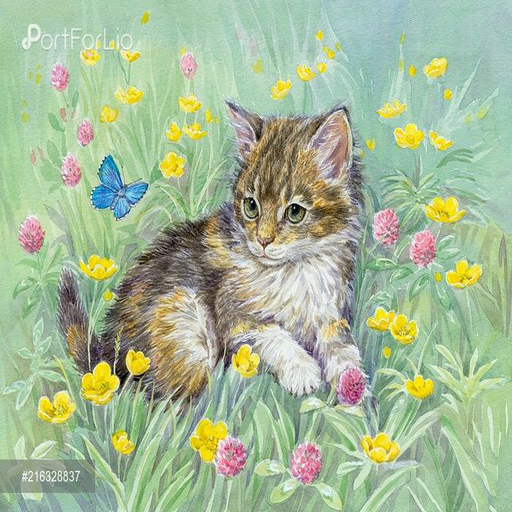}
    };
    \node[anchor=south, yshift=-2.5mm] at (FigA1.south) {\footnotesize Source};
\end{tikzpicture}\hspace{-1mm}%
% \hspace{-2.3mm}
% \raisebox{0.4in}{\rotatebox{90}{Target}}%
% \hspace{-1.3mm}
% 2nd Image
% \begin{tikzpicture}[x=1cm, y=1cm]
%     \node[anchor=south] (FigA2) at (0,0) {
%         \includegraphics[width=\imgwidth]{sec/X_supp/Fig/imgs/loss_ablation/dds.png}
%     };
%     \node[anchor=south, yshift=-2.5mm] at (FigA2.south) {\footnotesize w/o \ FPR};
% \end{tikzpicture}\hspace{-1mm}%
\begin{tikzpicture}[x=1cm, y=1cm]
    \node[anchor=south] (FigB1) at (0,0) {
        \includegraphics[width=\imgwidth]{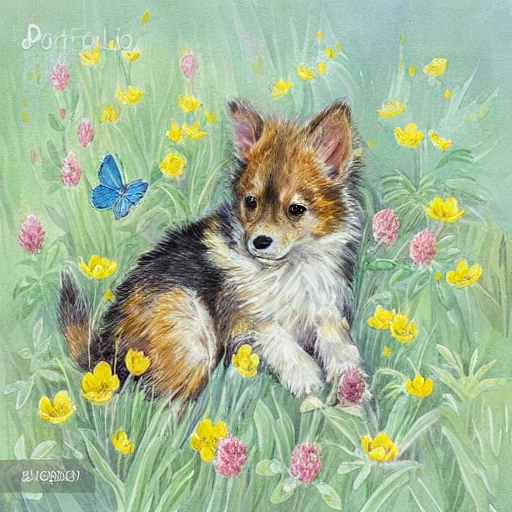}
    };
    \node[anchor=south, yshift=-2.5mm] at (FigB1.south) {\footnotesize Euclidean loss};
\end{tikzpicture}\hspace{-1mm}%
% % 3rd Image
\begin{tikzpicture}[x=1cm, y=1cm]
    \node[anchor=south] (FigC1) at (0,0) {
        \includegraphics[width=\imgwidth]{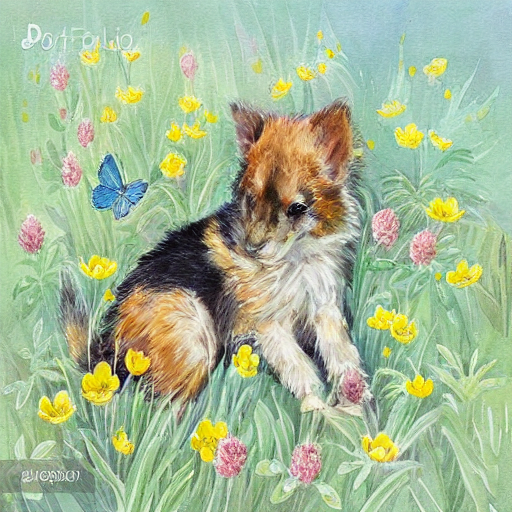}
    };
    \node[anchor=south, yshift=-2.5mm] at (FigC1.south) {\footnotesize L1 loss};
\end{tikzpicture}\hspace{-1mm}%
% 4th Image
\begin{tikzpicture}[x=1cm, y=1cm]
    \node[anchor=south] (FigD1) at (0,0) {
        \includegraphics[width=\imgwidth]{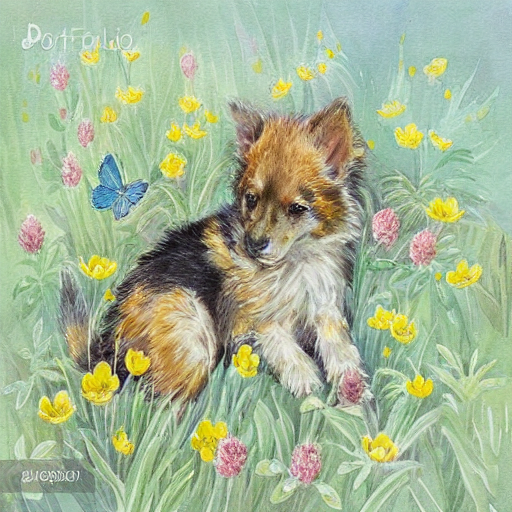}
    };
    \node[anchor=south, yshift=-2.5mm] at (FigD1.south) {\footnotesize SSIM loss};
\end{tikzpicture}

\vspace{-8pt}
\caption{\textbf{Ablation study for loss function.} Edited results of \textit{(first)} the source image from prompt \textit{``a drawing of a cat"} to \textit{``a drawing of a dog"} using \textit{(second)} Euclidean, \textit{(third)} L1, and \textit{(fourth)} SSIM loss function for FPR.}
\label{fig:sup_loss_abl}
\end{figure}

As defined in \cref{eq:fpr}
%Eq. (6)
, $d(\textbf{x}_1, \textbf{x}_2)$ can be any metric to calculate the difference between two inputs. For comparison, we
consider three different strategies: (1) Euclidean loss, (2) L1 loss, and (3) SSIM loss. 
As demonstrated in \cref{fig:sup_loss_abl}, all metrics can be applied to our method for image editing according to text prompts.
Among these, the use of Euclidean loss is particularly notable, as it effectively preserved the original information while producing visually superior results.

%%% [START] 6.1

\section{Implementation details}
\label{sec:s_implement}
% which model we used
% about baselines
% how to calculate background psnr & IoU
% calculated mask examples
For experiments, we implement our method based on the official code of CDS \footnote{\url{https://hyelinnam.github.io/CDS/}} by using Stable Diffusion v1.4. All baselines are implemented based on the official code and setting for each method.
For the proposed FPR, we set the scale $\lambda$ to $1.0$ and iteration $N$ to 3. 
The range of timesteps, optimization, learning rate, and number of optimization steps correspond to the default settings employed in DDS and CDS. 
% for experimental results in the main paper. 
All experiments are conducted on a single NVIDIA RTX 3090.

\section{Evaluation metrics}
\label{sec:s_evalmetric}
% which model we used
% about baselines
% how to calculate background psnr & IoU
% calculated mask examples
Our purpose is to preserve the source information by optimizing the score $\epsilon_\phi^\text{src}$. Thus, in addition to the LPIPS, we newly utilize IoU and background PSNR as our metrics to measure the structural similarity between the source and edited image.% commonly used in the field of image editing. 
%For LAION 5B dataset, we use intersection over union (IoU) of source and target mask. It represents how much the area of the prompt changes after image translation.
%And, in IP2P dataset \cite{brooks2023instructpix2pix}, we measure background PSNR because it is difficult to specify the modified part of the prompt as an object.
% IoU on LAION 5B dataset \cite{schuhmann2022laion} and background PSNR on . 
%in addition to the LPIPS commonly used in the field of image editing. 

\begin{figure}[H] % 1-column
\footnotesize
\centering 

% Adjust the image width to fit within one column
\newcommand{\imgwidth}{0.45\linewidth} % Set image width to 18% of the line width

\begin{subfigure}{0.49\linewidth}
\hspace{-2.3mm}
\raisebox{0.25in}{\rotatebox{90}{Images}}%
\hspace{-1.3mm}
% 1st Image
\begin{tikzpicture}[x=1cm, y=1cm]
    \node[anchor=south] (FigA1) at (0,0) {
        \includegraphics[width=\imgwidth]{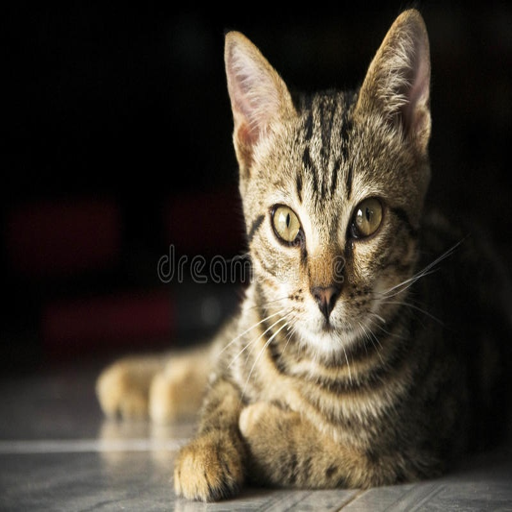}
    };
    % \node[anchor=south, yshift=-1mm] at (FigA1.south) {\footnotesize image};
\end{tikzpicture}\hspace{-1mm}%
% 2nd Image
\begin{tikzpicture}[x=1cm, y=1cm]
    \node[anchor=south] (FigB1) at (0,0) {
        \includegraphics[width=\imgwidth]{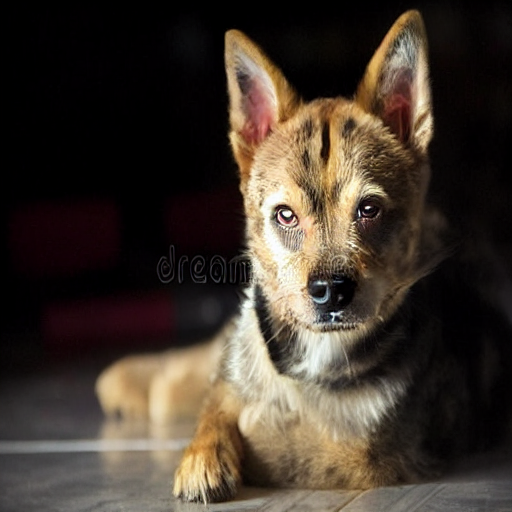}
    };
    % \node[anchor=south, yshift=-1mm] at (FigB1.south) {\footnotesize mask};
\end{tikzpicture}\hspace{-1mm}%

\vspace{6pt}

\hspace{-2.2mm}
\raisebox{0.35in}{\rotatebox{90}{Masks}}%
\hspace{-1.2mm}
% 1st Image
\begin{tikzpicture}[x=1cm, y=1cm]
    \node[anchor=south] (FigA2) at (0,0) {
        \includegraphics[width=\imgwidth]{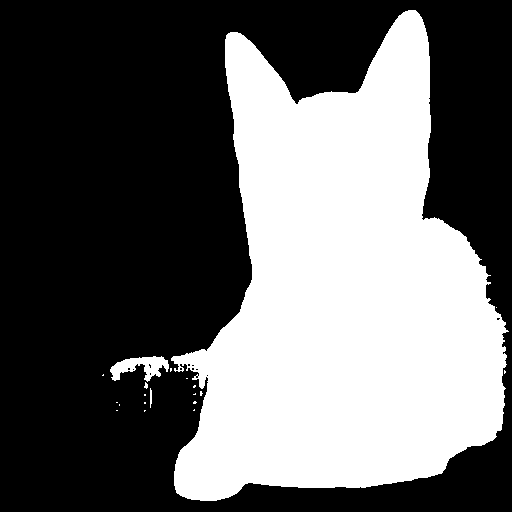}
    };
    \node[anchor=south, yshift=-2.5mm] at (FigA2.south) {\footnotesize source \textit{``cat"}};
\end{tikzpicture}\hspace{-1mm}%
% 2nd Image
\begin{tikzpicture}[x=1cm, y=1cm]
    \node[anchor=south] (FigB2) at (0,0) {
        \includegraphics[width=\imgwidth]{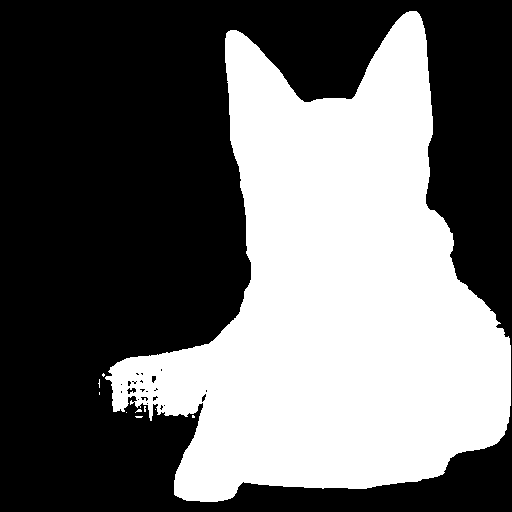}
    };
    \node[anchor=south, yshift=-2.5mm] at (FigB2.south) {\footnotesize target \textit{``dog"}};
\end{tikzpicture}\hspace{-1mm}%
\caption{IoU mask}
\end{subfigure}
\hfill
\begin{subfigure}{0.49\linewidth}
\hspace{-2.3mm}
\raisebox{0.35in}{\rotatebox{90}{Images}}%
\hspace{-1.3mm}
% 1st Image
\begin{tikzpicture}[x=1cm, y=1cm]
    \node[anchor=south] (FigA3) at (0,0) {
        \includegraphics[width=\imgwidth]{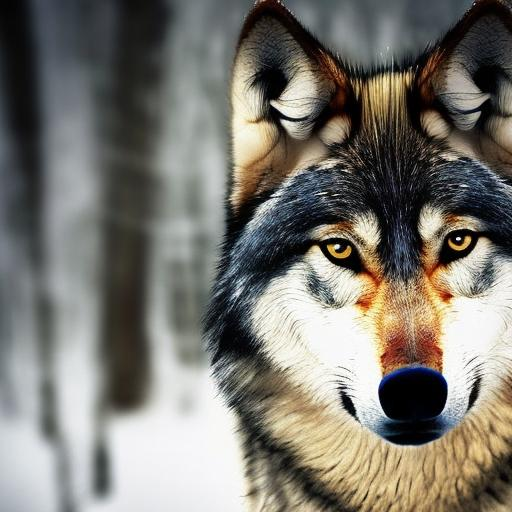}
    };
    \node[anchor=south, yshift=-2.5mm] at (FigA3.south) {\footnotesize source};
\end{tikzpicture}\hspace{-1mm}%
% 2nd Image
\begin{tikzpicture}[x=1cm, y=1cm]
    \node[anchor=south] (FigB3) at (0,0) {
        \includegraphics[width=\imgwidth]{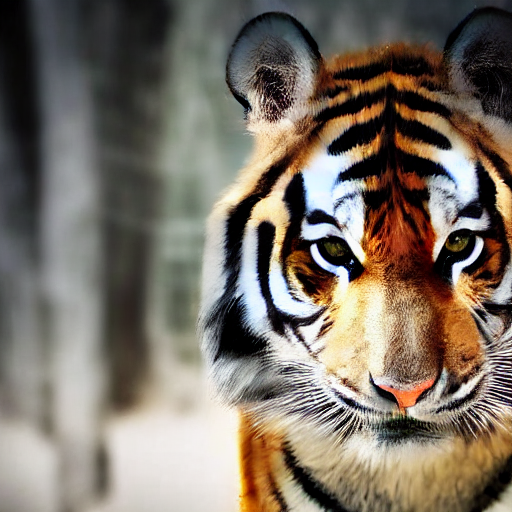}
    };
    \node[anchor=south, yshift=-2.5mm] at (FigB3.south) {\footnotesize target};
\end{tikzpicture}\hspace{-1mm}%

% \vspace{-3pt}

\hspace{-2.2mm}
\raisebox{0.35in}{\rotatebox{90}{Masks}}%
\hspace{-1.2mm}
% 1st Image
\begin{tikzpicture}[x=1cm, y=1cm]
    \node[anchor=south] (FigA4) at (0,0) {
        \includegraphics[width=\imgwidth]{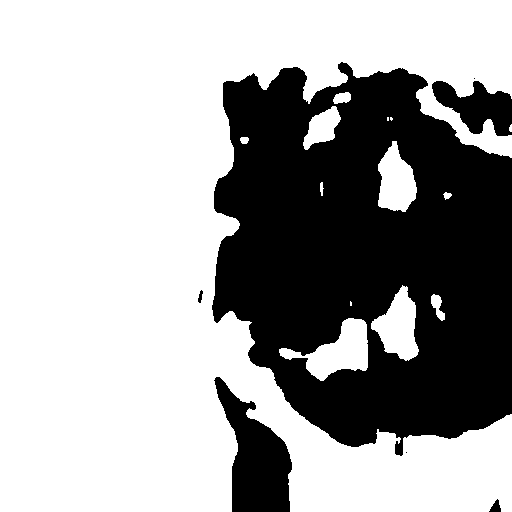}
    };
    \node[anchor=south, yshift=-2.5mm] at (FigA4.south) {\footnotesize mean};
\end{tikzpicture}\hspace{-1mm}%
% 2nd Image
\begin{tikzpicture}[x=1cm, y=1cm]
    \node[anchor=south] (FigB4) at (0,0) {
        \includegraphics[width=\imgwidth]{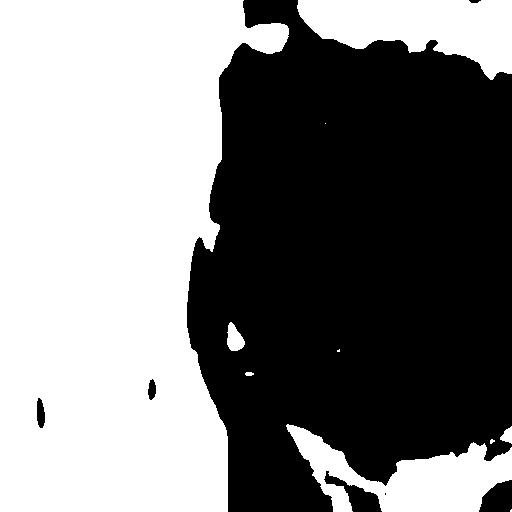}
    };
    \node[anchor=south, yshift=-2.5mm] at (FigB4.south) {\footnotesize median};
\end{tikzpicture}\hspace{-1mm}%
\caption{Background PSNR mask}
\end{subfigure}
\vspace{-5pt}
% Caption Text
\caption{\textbf{Calculated masks} for IoU and background PSNR. In (a), \textit{(second row)} each mask for \textit{(first row)} the source and target image is obtained by using lang-SAM for IoU. In (b), \textit{(second row)} a mask is calculated for \textit{(first row)} the source and target image to measure background PSNR between the masked source and target image. The mask can be generated by thresholding method, mean and median}
% \vspace{-5pt}
\label{fig:sup_mask}
\end{figure}

\noindent\textbf{IoU.} The aim of \textit{Cat-to-Others} task is to translate the cat into another animal. Thus, the segmentation mask of the cat and translated animal can be obtained using
%We calculate IoU for \textit{Cat-to-Others} task, We use 
the language Segment-Anything model (lang-SAM)\footnote{\url{https://github.com/paulguerrero/lang-sam}}, which is an open-source project to segment some objects from the text prompt. 
IoU of the source and target mask represents how much the area of the cat changes after image editing. The lower the IoU, the more similar the region of the cat and the region of the translated animal, meaning the overall shape is preserved.
To this end, first, the mask about the prompt is obtained from an image using lang-SAM. For example, `cat' is segmented from the source image to get the mask $M_{\text{src}}$, while `dog' is segmented from the edited image to obtain the mask $M_{\text{trg}}$, as shown in \cref{fig:sup_mask} (a). After getting masks, we calculate IoU from the masks that are given by:
\begin{equation*}
    \text{IoU}=\frac{\left(M_{\text{src}} \cap M_{\text{trg}}\right)}{\left(M_{\text{src}}\cup M_{\text{trg}}\right)}
\end{equation*}

\noindent\textbf{Background PSNR.} Since the editing prompts of IP2P dataset~\cite{brooks2023instructpix2pix} is complex than \textit{Cat-to-Others} dataset~\cite{schuhmann2022laion, nam2024contrastive}, it is hard to get mask by lang-SAM. Therefore, we use background PSNR to evaluate how much the original information is preserved. The residual of the source and target images is calculated, and the standard deviation $\sigma$ of each pixel of the residual image is computed with window size 30. Then, the mask $M_{\text{PSNR}}$ is acquired by thresholding the $\sigma$. Since the range of $\sigma$ varies according to the edited results for each method, we use the mean or median values of $\sigma$ to set an appropriate threshold (see \cref{fig:sup_mask} (b)). For the background PSNR of 
% \cref{tab:2Dquan}
Tab. 1, we use mean threshold. Finally, we calculate PSNR values from masked source and target images: 
\begin{equation*}
    \text{PSNR}_{\text{back}}=\text{PSNR}(M_{\text{PSNR}} \odot \mathbf{z}_{\text{src}}, M_{\text{PSNR}} \odot \mathbf{z}_{\text{trg}})
\end{equation*}
where $\odot$ is pixel-wise multiplication.
%%% [START] Ablation-fp_iter

\section{Extension to other methods}
\label{sec:s_extension}
% SDS image + FPR
% SDS NeRF + FPR
% CDS + FPR
% PDS + FPR
Since our method optimizes the source latent to estimate a more accurate score, it can be applied to other methods that are based on SDS despite that we report the results using our method to DDS.

During SDS optimization, FPR can be used to preserve the original content and reduce the blurry effect. As shown in \cref{fig:existingSDS}, the conserved rate of the information of the source image is controllable by the number of FPR iteration.

\begin{figure}[t] % 1-column
\footnotesize
\centering 

% Adjust the image width to fit within one column
\newcommand{\imgwidth}{1in} % Set image width to 16% of the line width

% 1st Image
\begin{tikzpicture}[x=1cm, y=1cm]
    \node[anchor=south] (FigA1) at (0,0) {
        \includegraphics[width=\imgwidth]{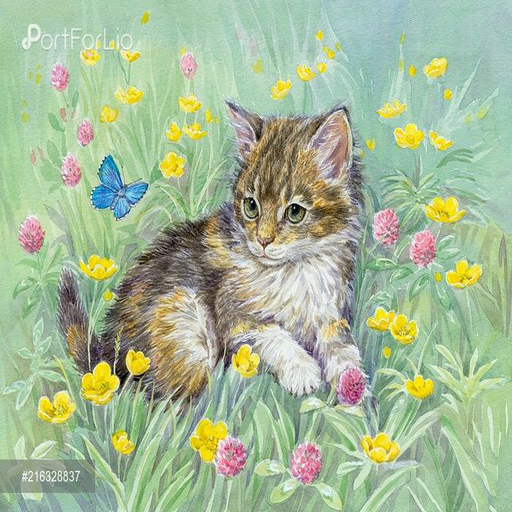}
    };
    \node[anchor=south, yshift=-1mm] at (FigA1.north) {\footnotesize Source};
\end{tikzpicture}\hspace{-1mm}%
% 2nd Image
\begin{tikzpicture}[x=1cm, y=1cm]
    \node[anchor=south] (FigB1) at (0,0) {
        \includegraphics[width=\imgwidth]{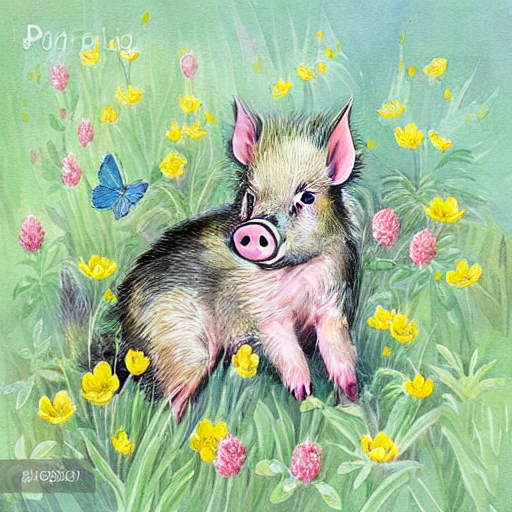}
    };
    \node[anchor=south, yshift=-1mm] at (FigB1.north) {\footnotesize CDS};
\end{tikzpicture}\hspace{-1mm}%
% % 3rd Image
\begin{tikzpicture}[x=1cm, y=1cm]
    \node[anchor=south] (FigC1) at (0,0) {
        \includegraphics[width=\imgwidth]{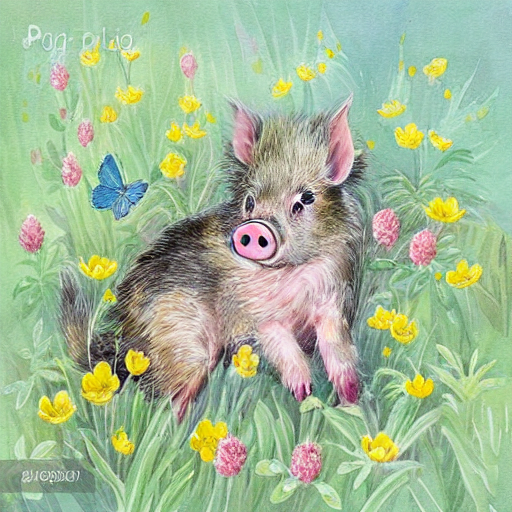}
    };
    \node[anchor=south, yshift=-1mm] at (FigC1.north) {\footnotesize FPR+CDS};
\end{tikzpicture}\hspace{-1mm}%
% 4th Image

\vspace{-8pt}
\caption{\textbf{CDS with FPR.} %regularization.} 
Given \textit{(first)} source image, source prompt \textit{``a drawing of a cat"}, and target prompt \textit{``a drawing of a pig"}, \textit{(second)} CDS translation, \textit{(third)} CDS optimization with FPR for $N=3$ and $\lambda=1.0$.}% Each result uses 200 steps for optimization.}
\label{fig:sup_fpr_cds}
\end{figure}

When the proposed FPR is integrated into CDS, the texture of the source image is further maintained, as illustrated in \cref{fig:sup_fpr_cds}. In addition, FPR promoted reducing the over-boosting of color often found in the translated images of CDS. This confirms that the proposed FPR can be a universal regularization to preserve the identity of the source image for text-guided image editing.
\input{Fig./Qual/Editing/existingSDS}
%%% [START] 6.1

% \paragraph{FPR + SDS in 2D} FPR can be applied to SDS optimization for a given source image and prompt to preserve the original contents and reduce the blurry effect. As shown in \cref{fig:existingSDS}, the conserved rate of the information of the source image is controllable by the number of iterations of FPR.
% \input{Fig./Qual/Editing/existingSDS}

\begin{figure}[!tbh]
\centering
\footnotesize

% 1st row
\raisebox{0.3in}{\rotatebox{90}{PDS}}%
\hspace{0.02mm}
\includegraphics[width=0.95\columnwidth]{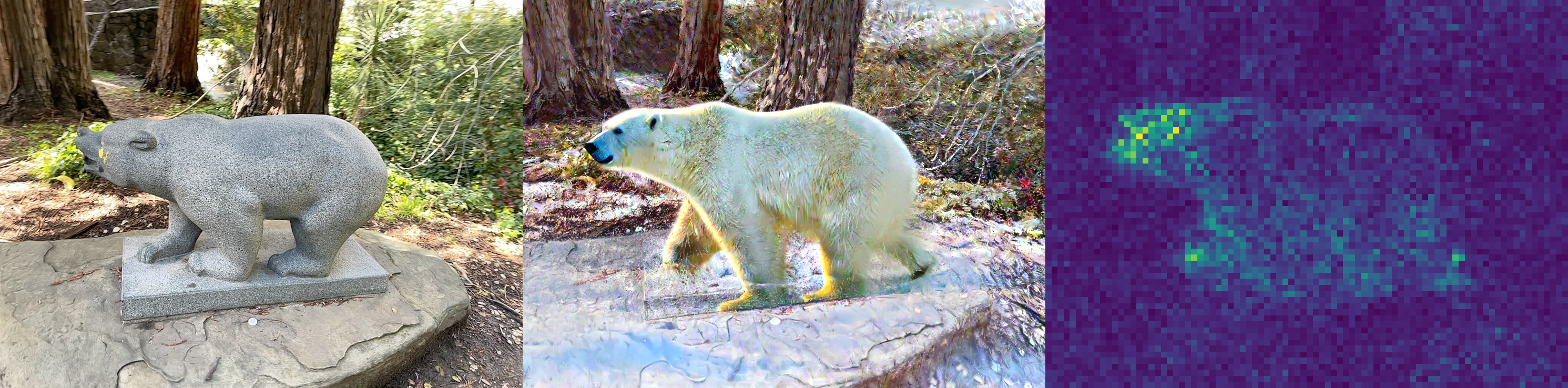}

\vspace{2pt}

% 2nd row
\raisebox{0.15in}{\rotatebox{90}{\textbf{FPR+PDS}}}%
\hspace{-0.1mm}
\includegraphics[width=0.95\columnwidth]{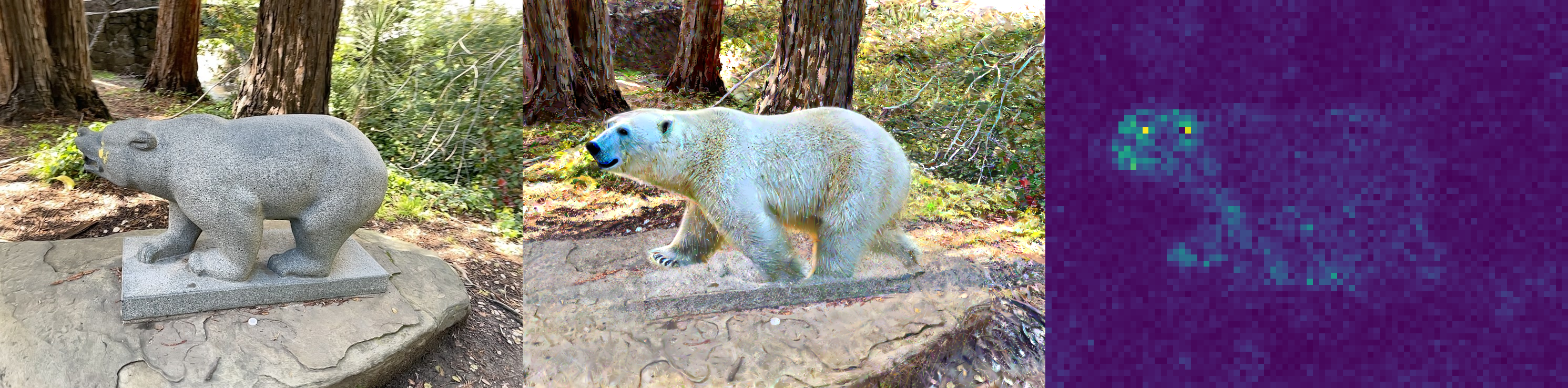}

\vspace{-2.2pt}
% text
\setulcolor{magenta}
\setul{0.3pt}{2pt}
\centering \textit{``A bear made of stone" $\to$ ``A \ul{polar} bear"} 
\vspace{-2.2pt}

\vspace{2pt}

% 3rd row
\raisebox{0.3in}{\rotatebox{90}{PDS}}%
\hspace{0.02mm}
\includegraphics[width=0.95\columnwidth]{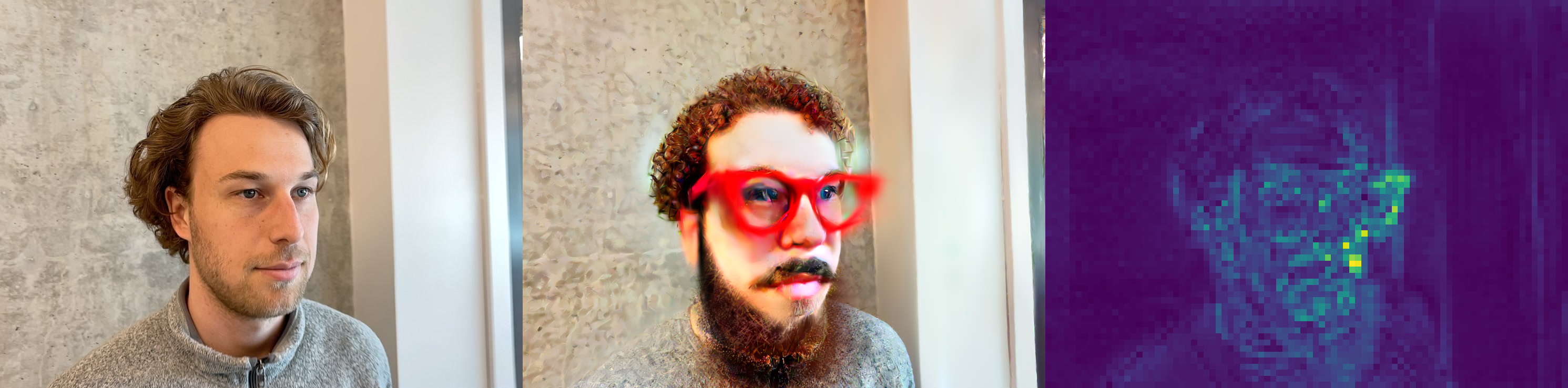}

\vspace{2pt}

% 4th row
\raisebox{0.15in}{\rotatebox{90}{\textbf{FPR+PDS}}}%
\hspace{-0.1mm}
\includegraphics[width=0.95\columnwidth]{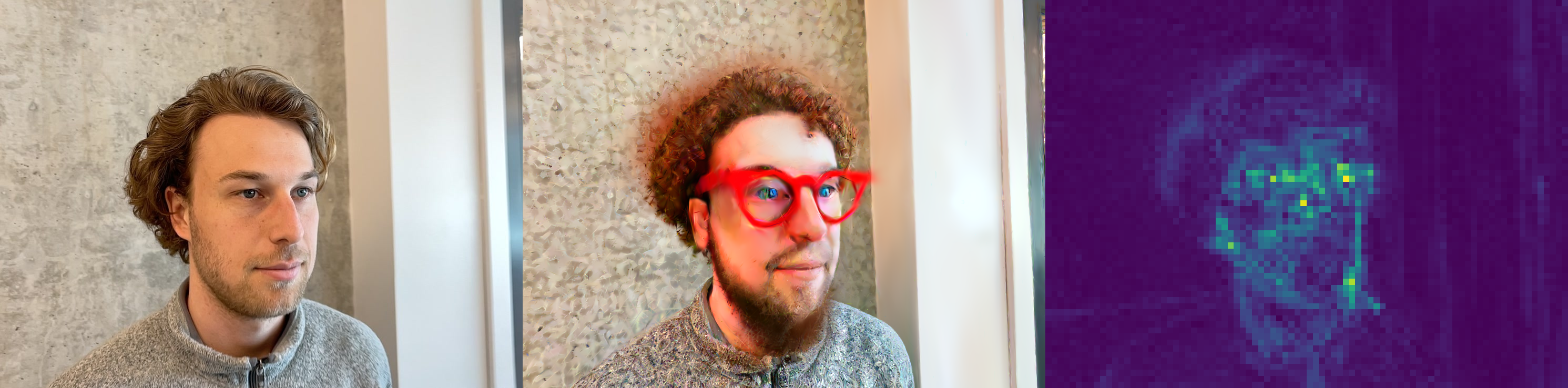}

\vspace{-2.2pt}
% text
\setulcolor{magenta}
\setul{0.3pt}{2pt}
\centering \textit{``A man with curly hair with a beard" \\ $\to$ ``A man \ul{wearing with red glasses} ... "} 
\vspace{-8pt}

\caption{\textbf{3D Qualitative results for PDS} on subset of Instruct-NeRf2NeRF \cite{haque2023instruct}. From left to right, each column represents the source image, the edited image, and the gradient weight. The gradient weight indicates which regions the model primarily references during the editing process. The results demonstrate that FPR operates effectively in End-to-End NeRF while preserving the structure and identity of the source image.}
\vspace{-8pt}
\label{fig:ids-vs-pds}
\end{figure}

\begin{figure}[!tbh] % 1-column
\centering
\footnotesize

% Adjust the image width so that 5 images fit within one column
\newcommand{\imgwidth}{0.18\linewidth}

% 1st Image
\begin{tikzpicture}[x=1cm, y=1cm]
    \node[anchor=south] (FigA1) at (0,0) {
        \includegraphics[width=\imgwidth]{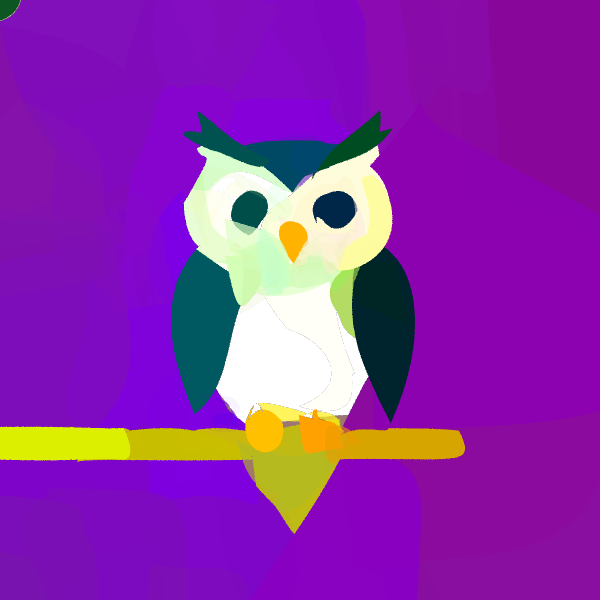}
    };
    \node[anchor=south, yshift=-1.5mm] at (FigA1.north) {\footnotesize Source};
\end{tikzpicture}\hspace{-1mm}%
% 2nd Image
\begin{tikzpicture}[x=1cm, y=1cm]
    \node[anchor=south] (FigB1) at (0,0) {
        \includegraphics[width=\imgwidth]{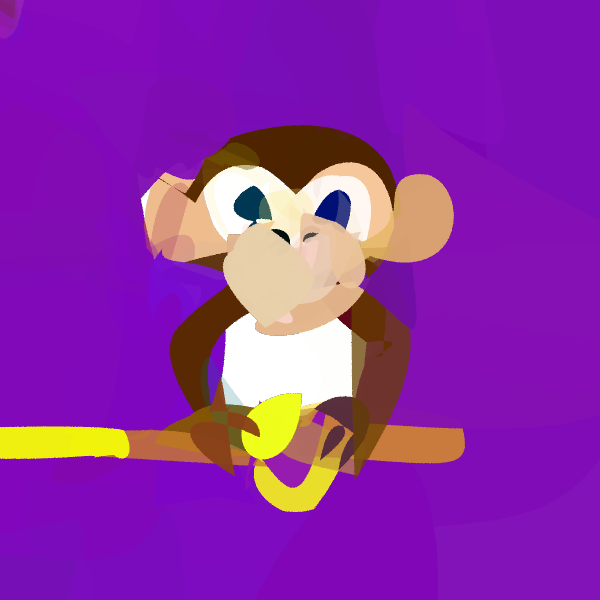}
    };
    \node[anchor=south, yshift=-1.5mm] at (FigB1.north) {\footnotesize \textbf{FPR+PDS}};
\end{tikzpicture}\hspace{-1mm}%
% 3rd Image
\begin{tikzpicture}[x=1cm, y=1cm]
    \node[anchor=south] (FigC1) at (0,0) {
        \includegraphics[width=\imgwidth]{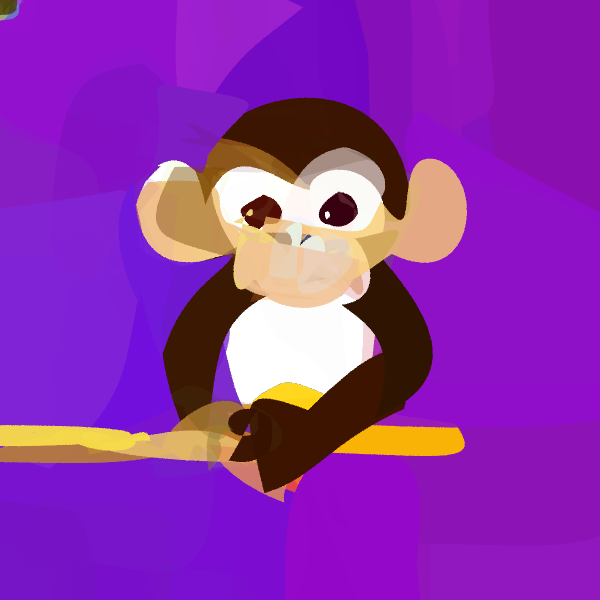}
    };
    \node[anchor=south, yshift=-1.5mm] at (FigC1.north) {\footnotesize PDS};
\end{tikzpicture}\hspace{-1mm}%
% 4th Image
\begin{tikzpicture}[x=1cm, y=1cm]
    \node[anchor=south] (FigD1) at (0,0) {
        \includegraphics[width=\imgwidth]{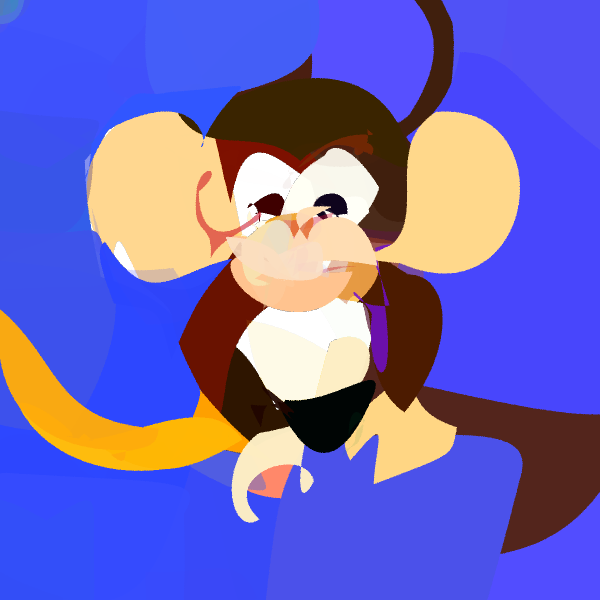}
    };
    \node[anchor=south, yshift=-1.5mm] at (FigD1.north) {\footnotesize DDS};
\end{tikzpicture}\hspace{-1mm}%
% 5th Image
\begin{tikzpicture}[x=1cm, y=1cm]
    \node[anchor=south] (FigE1) at (0,0) {
        \includegraphics[width=\imgwidth]{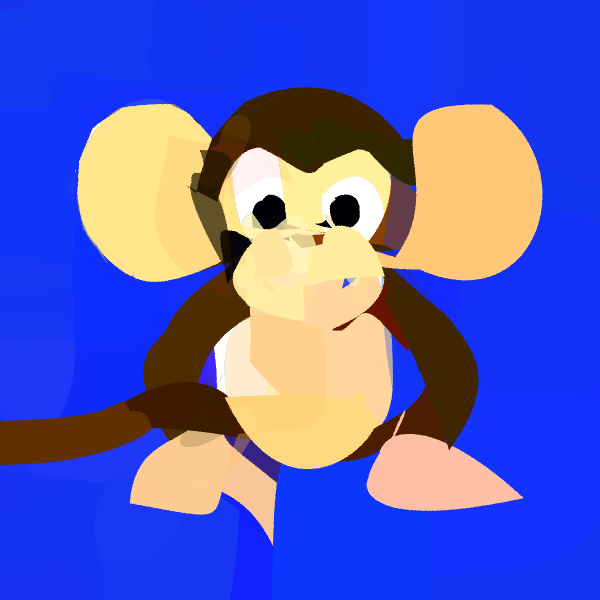}
    };
    \node[anchor=south, yshift=-1.5mm] at (FigE1.north) {\footnotesize SDS};
\end{tikzpicture}

\vspace{-5pt}
\setulcolor{magenta}
\setul{0.3pt}{2pt}
\centering
\textbf{SVG:} \textit{``An owl" $\to$ ``\ul{A monkey}"} 
\vspace{-8pt}
\caption{\textbf{2D Qualitative results for PDS on VectorFusion \cite{jain2023vectorfusion}}. In SVG editing, our method can be utilized with PDS and help the source identity maintained.}
% \vspace{-8pt}
\label{fig:pds_svg}
\end{figure}

Furthermore, FPR can help optimize not only pixel space but also the parametric editor such as PDS~\cite{koo2024posterior}. As demonstrated in \cref{fig:ids-vs-pds}, \cref{fig:pds_svg}, and \cref{tab:nerf_svg}, the edited results with our method show that FPR assists in maintaining the original contents. By comparing the first and second rows of \cref{fig:ids-vs-pds}, the use of FPR results in the preservation of source components more effectively compared to PDS. Similarly, in the third and fourth rows, the results obtained using FPR retain key original features, such as the shape and color of the face as well as the color of the clothing. Furthermore, the gradient weights, FPR assigns minimal weight to the structure of the source image, such as the background, while primarily focusing the weights on the editing points. For 3D and 2D editing, we implement the experiments based on official code of PDS\footnote{\url{https://github.com/KAIST-Visual-AI-Group/PDS}}. We use the subset of Instruct-NeRF2NeRF \cite{haque2023instruct} for 3D editing and Scalable Vector Graphics (SVGs) with their text description used in \cite{jain2023vectorfusion} for 2D editing. 
% \paragraph{FPR + SDS in 3D.}

% \paragraph{FPR + CDS} As shown in \cref{fig:sup_fpr_cds}, FPR can regulate the  control the 
% As shown in CDS's translation of \cref{fig:sup_fpr_cds}, CDS has the effect of boosting color. FPR makes the translated result of CDS 
% \input{sec/X_supp/Fig/comp/fpr_cds}
% \paragraph{FPR + PDS.}
%%% [START] end-to-end Nerf

% \begin{table}[ht]
%     \centering
\begin{table}[ht]
    \centering
    % 표 안의 글자 크기를 한 단계 줄임
    \small
    % 열 간 간격을 기본보다 줄임 (기본 약 6pt)
    \setlength{\tabcolsep}{1pt}
    % 행 높이 계수를 줄임 (기본 1.0)
    \renewcommand{\arraystretch}{0.5}
    \begin{tabular}{lcc cc}
    \toprule
          & \multicolumn{2}{c}{\scriptsize\textbf{NeRF}} & \multicolumn{2}{c}{\scriptsize\textbf{SVG}} \\
          \cmidrule(lr){2-3} \cmidrule(lr){4-5}
    Metric & {\scriptsize CLIP($\uparrow$)} & {\scriptsize LPIPS($\downarrow$)} & {\scriptsize CLIP($\uparrow$)} & {\scriptsize LPIPS($\downarrow$)} \\
    \midrule
    SDS      & 0.305          & 0.814          & \textbf{0.346} & 0.552 \\
    DDS      & \textbf{0.306} & 0.875          & 0.344          & 0.557 \\
    PDS      & 0.292          & 0.662          & 0.324          & 0.326 \\
    \textbf{Ours+PDS} & 0.295          & \textbf{0.587} & 0.327          & \textbf{0.274} \\
    \bottomrule
    \end{tabular}
    \caption{\textbf{Quantitative results} for PDS.}
    \label{tab:nerf_svg}
\end{table}

\section{Additional results}
\label{sec:s_addresults}
% various prompts for one image
% cat2others qual quan
%%% [START] NeRF Synthetic data Results 
\begin{figure}[!tbh] % 2-column
\footnotesize
\centering 

% 1st row
\hspace{-2.3mm}
\raisebox{0.35in}{\rotatebox{90}{Cow}}%
\hspace{-1.3mm}
\hspace{0mm}\begin{tikzpicture}[x=3.5cm, y=3.5cm]
\node[anchor=south] (FigA) at (0,0) {\includegraphics[width=0.78in]{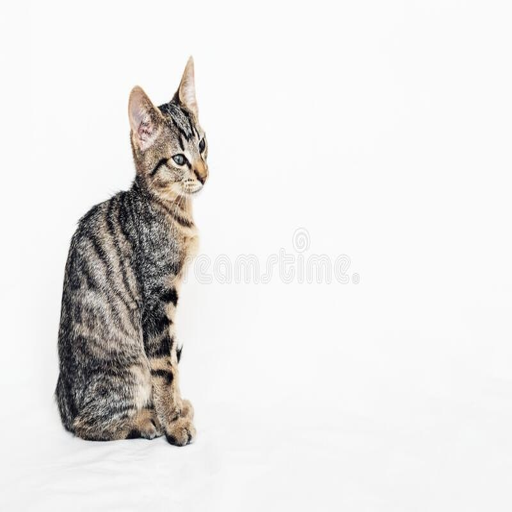}};
\node[anchor=south, yshift=0mm] at (FigA.north) {\footnotesize Source};
\end{tikzpicture}\hspace{-1.1mm}%
\begin{tikzpicture}[x=3.5cm, y=3.5cm]
\node[anchor=south] (FigB) at (0,0) {\includegraphics[width=0.78in]{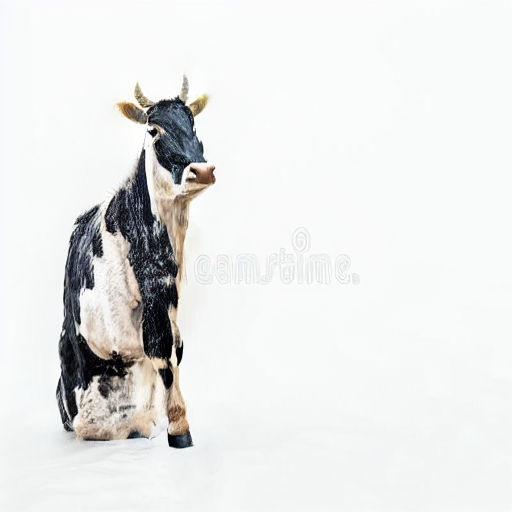}};
\node[anchor=south, yshift=0mm] at (FigB.north) {\footnotesize \textbf{IDS (Ours)}};
\end{tikzpicture}\hspace{-1.1mm}%
\begin{tikzpicture}[x=3.5cm, y=3.5cm]
\node[anchor=south] (FigC) at (0,0) {\includegraphics[width=0.78in]{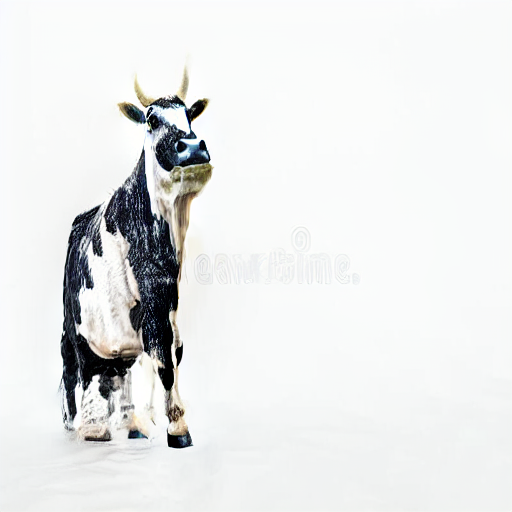}};
\node[anchor=south, yshift=0mm] at (FigC.north) {\footnotesize CDS};
\end{tikzpicture}\hspace{-1.1mm}%
\begin{tikzpicture}[x=3.5cm, y=3.5cm]
\node[anchor=south] (FigD) at (0,0) {\includegraphics[width=0.78in]{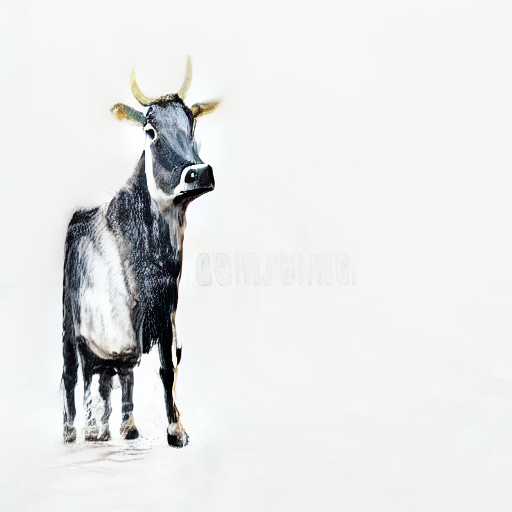}};
\node[anchor=south, yshift=0mm] at (FigD.north) {\footnotesize DDS};
\end{tikzpicture}%
\vspace{-3pt}

% 2nd row
\hspace{-2.5mm}
\raisebox{0.35in}{\rotatebox{90}{Dog}}%
\hspace{-1.5mm}
\hspace{0mm}\begin{tikzpicture}[x=3.5cm, y=3.5cm]
\node[anchor=south] (FigA2) at (0,0) {\includegraphics[width=0.78in]{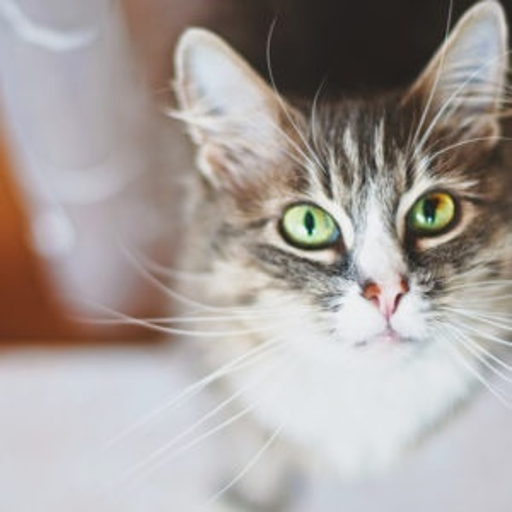}};
\end{tikzpicture}\hspace{-1.1mm}%
\begin{tikzpicture}[x=3.5cm, y=3.5cm]
\node[anchor=south] (FigB2) at (0,0) {\includegraphics[width=0.78in]{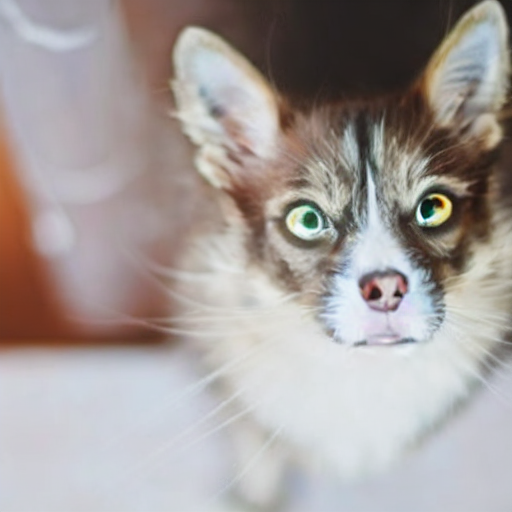}};
\end{tikzpicture}\hspace{-1.1mm}%
\begin{tikzpicture}[x=3.5cm, y=3.5cm]
\node[anchor=south] (FigC2) at (0,0) {\includegraphics[width=0.78in]{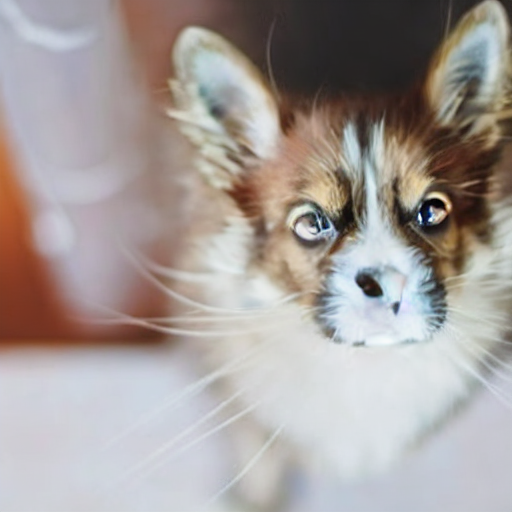}};
\end{tikzpicture}\hspace{-1.1mm}%
\begin{tikzpicture}[x=3.5cm, y=3.5cm]
\node[anchor=south] (FigE2) at (0,0) {\includegraphics[width=0.78in]{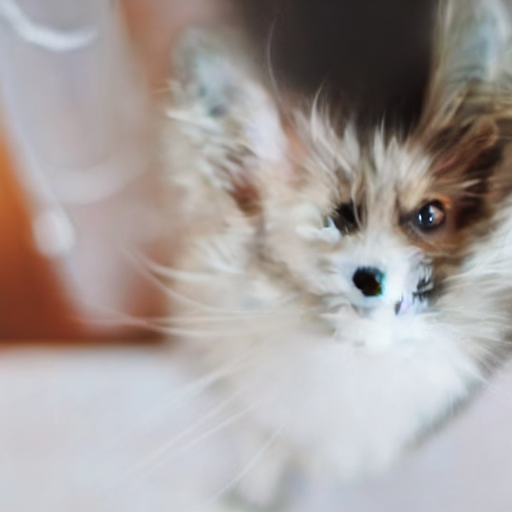}};
\end{tikzpicture}

\vspace{-3pt}

% 3rd row
\hspace{-2.3mm}
\raisebox{0.35in}{\rotatebox{90}{Lion}}%
\hspace{-1.3mm}
\hspace{0mm}\begin{tikzpicture}[x=3.5cm, y=3.5cm]
\node[anchor=south] (FigA4) at (0,0) {\includegraphics[width=0.78in]{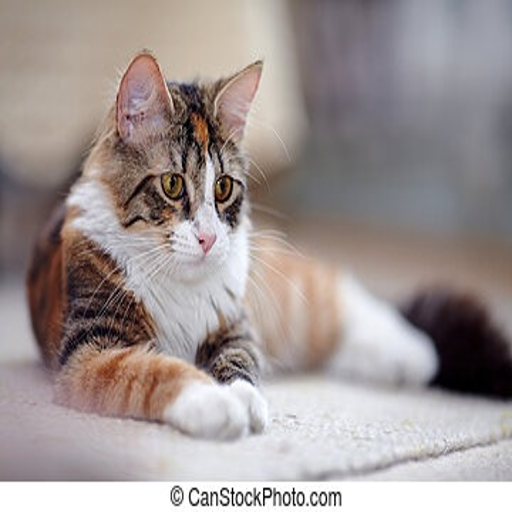}};
\end{tikzpicture}\hspace{-1.1mm}%
\begin{tikzpicture}[x=3.5cm, y=3.5cm]
\node[anchor=south] (FigB4) at (0,0) {\includegraphics[width=0.78in]{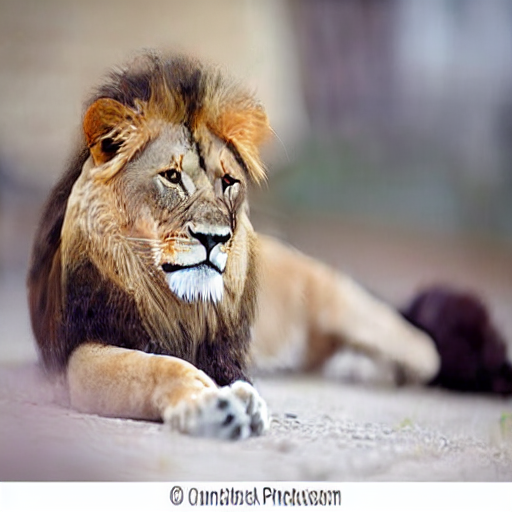}};
\end{tikzpicture}\hspace{-1.1mm}%
\begin{tikzpicture}[x=3.5cm, y=3.5cm]
\node[anchor=south] (FigC4) at (0,0) {\includegraphics[width=0.78in]{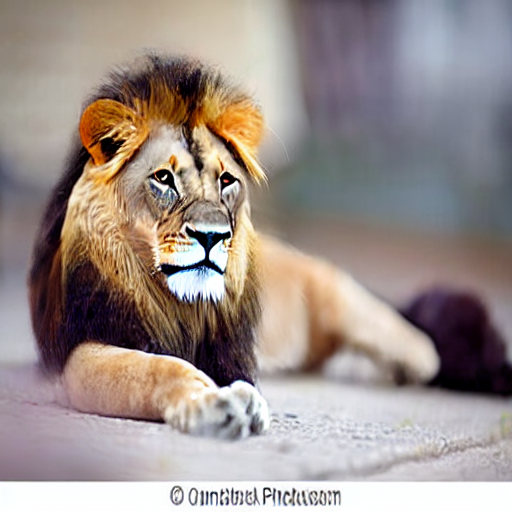}};
\end{tikzpicture}\hspace{-1.1mm}%
\begin{tikzpicture}[x=3.5cm, y=3.5cm]
\node[anchor=south] (FigE4) at (0,0) {\includegraphics[width=0.78in]{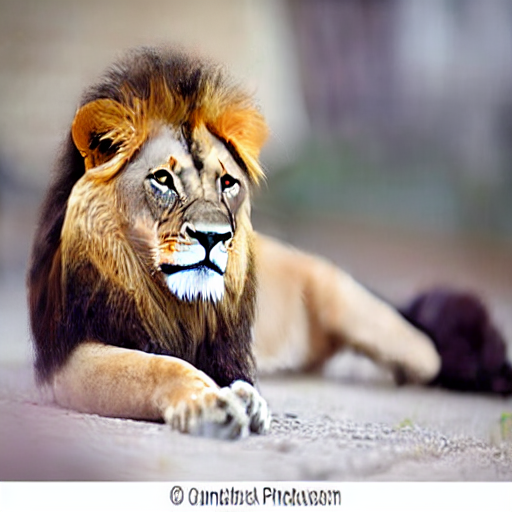}};
\end{tikzpicture}

\vspace{-3pt}

% 4th row
\hspace{-2.5mm}
\raisebox{0.25in}{\rotatebox{90}{Squirrel}}%
\hspace{-1.5mm}
\hspace{0mm}\begin{tikzpicture}[x=3.5cm, y=3.5cm]
\node[anchor=south] (FigA5) at (0,0) {\includegraphics[width=0.78in]{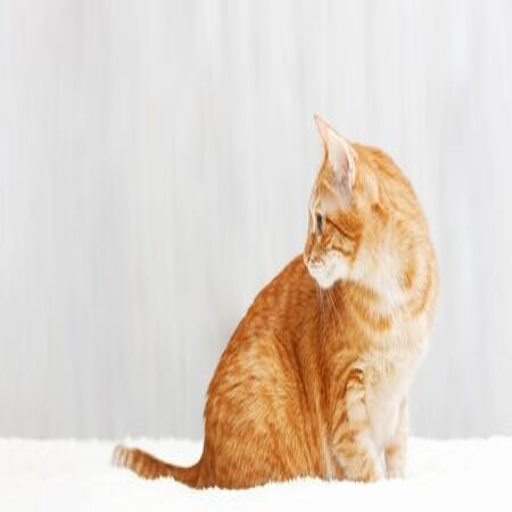}};
\end{tikzpicture}\hspace{-1.1mm}%
\begin{tikzpicture}[x=3.5cm, y=3.5cm]
\node[anchor=south] (FigB5) at (0,0) {\includegraphics[width=0.78in]{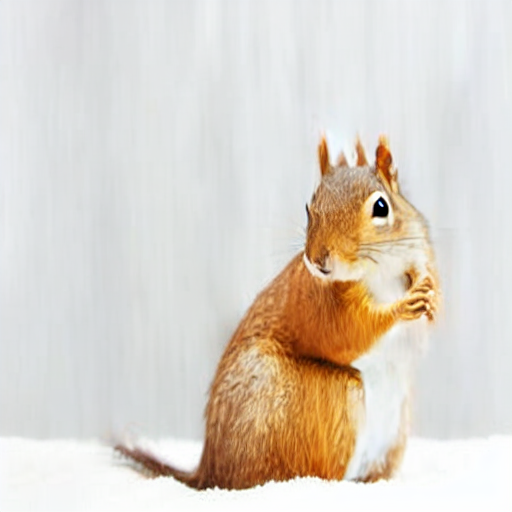}};
\end{tikzpicture}\hspace{-1.1mm}%
\begin{tikzpicture}[x=3.5cm, y=3.5cm]
\node[anchor=south] (FigC5) at (0,0) {\includegraphics[width=0.78in]{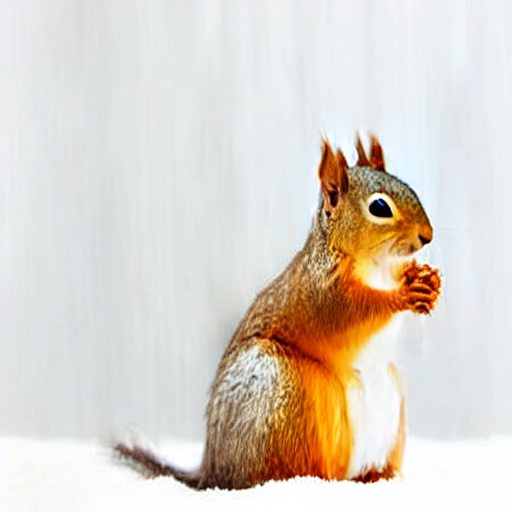}};
\end{tikzpicture}\hspace{-1.1mm}%
\begin{tikzpicture}[x=3.5cm, y=3.5cm]
\node[anchor=south] (FigE5) at (0,0) {\includegraphics[width=0.78in]{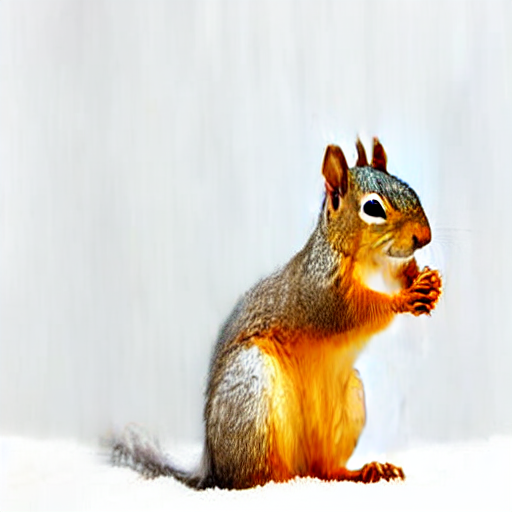}};
\end{tikzpicture}

\vspace{-3pt}

% 5th row
\hspace{-2.5mm}
\raisebox{0.35in}{\rotatebox{90}{Tiger}}%
\hspace{-1.5mm}
\hspace{0mm}\begin{tikzpicture}[x=3.5cm, y=3.5cm]
\node[anchor=south] (FigA4) at (0,0) {\includegraphics[width=0.78in]{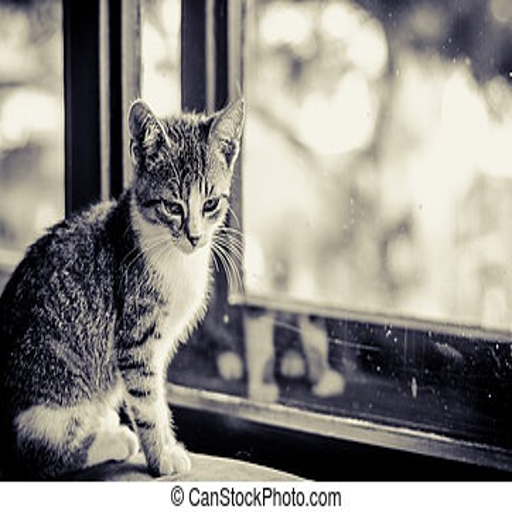}};
\end{tikzpicture}\hspace{-1.1mm}%
\begin{tikzpicture}[x=3.5cm, y=3.5cm]
\node[anchor=south] (FigB4) at (0,0) {\includegraphics[width=0.78in]{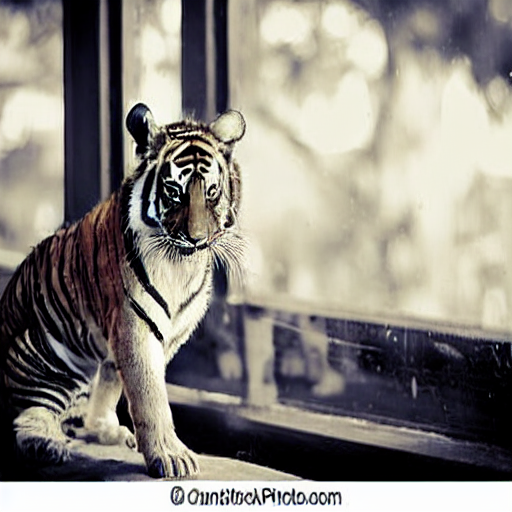}};
\end{tikzpicture}\hspace{-1.1mm}%
\begin{tikzpicture}[x=3.5cm, y=3.5cm]
\node[anchor=south] (FigC4) at (0,0) {\includegraphics[width=0.78in]{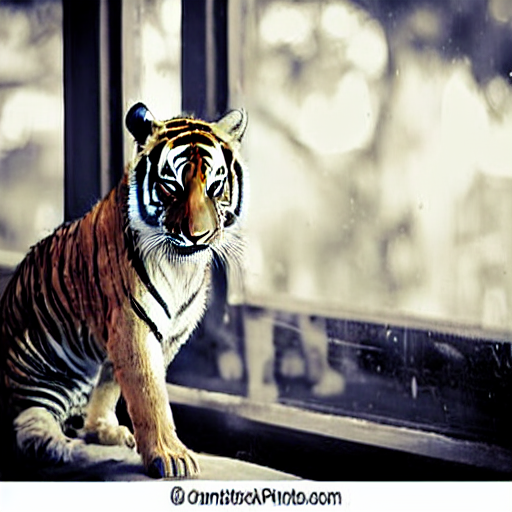}};
\end{tikzpicture}\hspace{-1.1mm}%
\begin{tikzpicture}[x=3.5cm, y=3.5cm]
\node[anchor=south] (FigE4) at (0,0) {\includegraphics[width=0.78in]{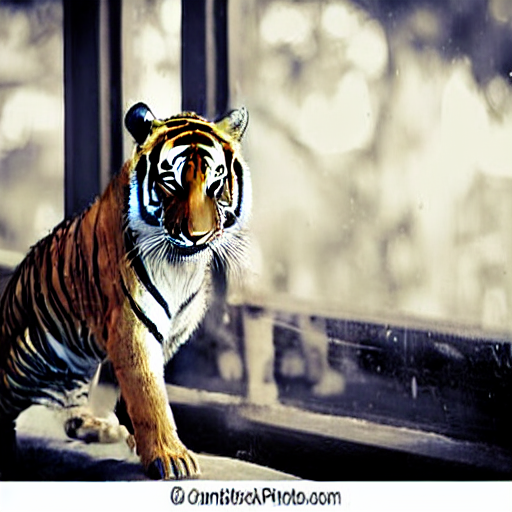}};
\end{tikzpicture}

\vspace{-5pt}

\caption{\textbf{Qualitative results} of \textit{Cat-to-Others} task. The leftmost text means each target prompt, and each row shows the editing results from `Cat" to the target prompt.}
\label{fig:sup_cat2others}
\end{figure}
\vspace{-3pt}

We also provide qualitative results for \textit{Cat-to-Others} task, as demonstrated in \cref{fig:sup_cat2others}. With DDS and CDS, the direction of the gaze changes when translated from the cat to the squirrel, while it remains the same with IDS.
Note that the proposed IDS can also retain the hue of the source image without overemphasizing the colors, as demonstrated in \textit{Cat-to-Tiger} task. This confirms that the proposed IDS consistently offers suitable editing of cat images into the diverse animals, while conserving the identity of the source against other algorithms. 

The trends in the quantitative results are also consistent with the qualitative result, as represented in \cref{tab:sup_cat2others}.
%In addition to the results in \cref{sec:5.1},
Our method provides the best performance for LPIPS and IoU in most \textit{Cat-to-Others} tasks. This shows again that the self-correction of the score using the proposed algorithm is crucial for maintaining the identity.
%as demonstrated in \cref{tab:sup_cat2others}. Furthermore, as discussed in the main paper, IDS maintains the identity of the source image such as pose and color (see \cref{fig:sup_cat2others}).

\begin{table*}[!thb]
\centering
\resizebox{0.95\textwidth}{!}{
\small{
\begin{tabular}{c|cc|cc|cc|cc|cc}
\hline
& \multicolumn{2}{c|}{cat2cow} & \multicolumn{2}{c|}{cat2dog} & \multicolumn{2}{c|}{cat2lion} & \multicolumn{2}{c|}{cat2tiger} & \multicolumn{2}{c}{cat2penguin} \\ 
\hline
\multicolumn{1}{c|}{Metric} & LPIPS ($\downarrow$) & IoU ($\uparrow$) & LPIPS ($\downarrow$) & IoU ($\uparrow$) & LPIPS ($\downarrow$) & IoU ($\uparrow$) & LPIPS ($\downarrow$) & IoU ($\uparrow$) & LPIPS ($\downarrow$) & IoU ($\uparrow$) \\ 
\hline
P2P \cite{hertzprompt}& 0.43 & 0.57 & 0.42 & 0.51 & 0.46 & 0.57 & 0.47 & 0.57 & 0.46 & 0.54 \\
PnP \cite{tumanyan2023plug}& 0.52 & 0.55 & 0.47 & 0.59 & 0.51 & 0.58 & 0.52 & 0.58 & 0.52 & 0.52 \\
DDS \cite{hertz2023delta}& 0.29 & 0.65 & 0.22 & 0.72 & 0.29 & 0.69 & 0.30 & 0.71 & 0.28 & 0.66 \\  
CDS \cite{nam2024contrastive}& 0.25 & 0.72 & 0.19 & 0.74 & 0.25 & \textbf{0.74} & 0.27 & 0.75 & 0.24 & \textbf{0.72} \\
\hline
\textbf{IDS (Ours)}& \textbf{0.21} & \textbf{0.74} & \textbf{0.17} & \textbf{0.75} & \textbf{0.21} & 0.71 & \textbf{0.21} & \textbf{0.76} & \textbf{0.21} & \textbf{0.72} \\
\hline
\end{tabular}
}
}
\vspace{-5pt}
\caption{\textbf{Quantitative results} for \textit{Cat-to-Others} task. LPIPS \cite{zhang2018unreasonable} and IoU are used. Lower LPIPS and higher IoU mean better identity preserving.}
\label{tab:sup_cat2others}
\end{table*}

%P2P \cite{hertzprompt}& 0.5798 & 0.4229 & 0.5184 & 0.4605 & 20.88 & 0.4695 \\
%PnP \cite{tumanyan2023plug}& 0.5507 & 0.5191 & ??? & 0.5245 & 23.81 & 0.3882 \\
%DDS \cite{hertz2023delta}& 0.6897 & 0.2838 & 0.6456 & 0.2996 & 26.02 & 0.2398 \\  
%CDS \cite{nam2024contrastive}& 0.7249 & 0.2485 & 0.7054 & 0.2612 & 27.35 & 0.2099 \\

%\input{sec/X_supp/s5_AblationStudies}

\section{Limitations}
\label{sec:supp_limit}
\noindent\textbf{Success rate.} As discussed in \cref{sec:limit}, %Sec. 7, 
our method optimizes the latents only for source information, resulting in low CLIP scores. To demonstrate it does not mean \textit{"IDS fails to translate the source image"}, we measure the success rate. To calculate the success rate, we classify the transformed images with the pre-trained ResNet classifier on the \textit{Cat-to-dog} task. We treat the results of classified top 1 as the success sample if they belonged to the class of \textit{dog}. The success rate in \cref{tab:success} claims that the low CLIP score of IDS did not fall to convert, but occurred in the process of maintaining the source identity.

% EDIT
\noindent\textbf{Failure case for complex prompt.} Because our method only considers the source information, it struggles with translating the given image for complex text prompts. Although we tried to modify the image with more complex prompts, it failed not only in our method but also in all SDS-based translation methods, as shown in \cref{fig:complex_ex}.
% Fig. 8.
% we try to modify the image with more complex prompts. Although it failed 

\noindent\textbf{Computational overhead.} Our method requires additional computational costs due to repetitive adaptations of FPR for each optimization steps. However, it can be controlled by adjusting hyperparameters such as the number of FPR iterations or the number of optimization steps, as reported in \cref{tab:overhead}. %Tab. 4.

\begin{table}[h]
% \scriptsize
\centering
\begin{tabular}{c|c|c|c}
\hline
& IDS & CDS & DDS  \\ 
\hline
success rate (\%) & 37.40 & 34.87 & 34.03 \\
\hline
\end{tabular}
\vspace{-5pt}
\caption{\textbf{Success rate} for \textit{Cat-to-dog} task. A higher score means more translated results are classified as \textit{dog}.}
\label{tab:success}
\end{table}

\section{Social impact}
\label{sec:s_social}
% simple description
By optimizing for a given image, our method properly mitigates the undesired biases introduced by the generative priors of large text-to-image diffusion models. However, the issue of bias toward target information persists. Furthermore, the method's potential misuse for generating fake content highlights a critical ethical challenge commonly associated with image editing techniques. To mitigate these risks, it is essential to implement robust safeguards, such as stricter content authentication mechanisms, and ethical guidelines for usage.

% {
%     \small
%     \bibliographystyle{ieeenat_fullname}
%     \bibliography{main}
% }


\end{document}